\pdfoutput=1
\documentclass[journal]{IEEEtran}

\usepackage{amsmath,amsfonts,amssymb}
\usepackage{array}
\usepackage{textcomp}
\usepackage{stfloats}
\usepackage{url}
\usepackage{graphicx}
\usepackage{cite}
\usepackage{booktabs}
\usepackage{multirow}
\usepackage{float}
\usepackage{pgfplots}
\pgfplotsset{compat=1.17}
\usetikzlibrary{shapes.geometric, arrows.meta, positioning, fit, backgrounds, calc, decorations.pathreplacing, patterns, chains, matrix}

\newcommand{\history}[1]{}
\newcommand{\doi}[1]{}
\newcommand{\tfootnote}[1]{}
\newcommand{\corresp}[1]{}

\newcommand{\EOD}{}

\newcommand{\vect}[1]{\mathbf{#1}}
\newcommand{\mat}[1]{\mathbf{#1}}

\begin{document}

\title{\textit{\small This work has been submitted to Applied Intelligence (Springer) for possible publication.}\\[0.5em]Adaptive Regime-Aware Stock Price Prediction Using Autoencoder-Gated Dual Node Transformers with Reinforcement Learning Control}
\author{\IEEEauthorblockN{Mohammad Al Ridhawi, Mahtab Haj Ali, and Hussein Al Osman}\\
\IEEEauthorblockA{School of Electrical Engineering and Computer Science,\\University of Ottawa, Ottawa, Canada\\e-mail: malri039@uottawa.ca}}

\markboth
{}
{}

\maketitle

\begin{abstract}
Stock markets exhibit regime-dependent behavior where prediction models optimized for stable conditions often fail during volatile periods. Existing approaches typically treat all market states uniformly or require manual regime labeling, which is expensive and quickly becomes stale as market dynamics evolve. This paper introduces an adaptive prediction framework that adaptively identifies deviations from normal market conditions and routes data through specialized prediction pathways. The architecture consists of three components: (1) an autoencoder trained on normal market conditions that identifies anomalous regimes through reconstruction error, (2) dual node transformer networks specialized for stable and event-driven market conditions respectively, and (3) a Soft Actor-Critic reinforcement learning controller that adaptively tunes the regime detection threshold and pathway blending weights based on prediction performance feedback. The reinforcement learning component enables the system to learn adaptive regime boundaries, defining anomalies as market states where standard prediction approaches fail. Experiments on 20 S\&P 500 stocks spanning 1982 to 2025 demonstrate that the proposed framework achieves 0.68\% mean absolute percentage error (MAPE) for one-day predictions without the reinforcement controller and 0.59\% MAPE with the full adaptive system, compared to 0.80\% for the baseline integrated node transformer. Directional accuracy reaches 72\% with the complete framework. The system maintains robust performance during high-volatility periods, with MAPE below 0.85\% when baseline models exceed 1.5\%. Ablation studies confirm that each component contributes meaningfully: autoencoder routing accounts for 36\% relative MAPE degradation upon removal, followed by the SAC controller at 15\% and the dual-path architecture at 7\%.
\end{abstract}


%========================================
% SECTION 1: INTRODUCTION
%========================================
\section{Introduction}
\label{sec:introduction}

\IEEEPARstart{F}{inancial} markets operate across distinct regimes characterized by different statistical properties, volatility levels, and correlation structures \cite{hamilton1989new}. During stable periods, price movements follow relatively predictable patterns driven by fundamental factors and gradual information incorporation. Crisis periods, earnings announcements, and macroeconomic shocks induce abrupt shifts in market behavior where historical patterns provide limited guidance. Models trained on aggregate historical data often perform well on average but degrade under volatile or event-driven conditions, where robust prediction is especially important.

Prior work on stock prediction has often treated market conditions as homogeneous. Graph neural networks capture cross-sectional dependencies \cite{chen2021graph}, transformers model temporal dynamics \cite{vaswani2017attention}, and sentiment analysis incorporates qualitative signals \cite{devlin2019bert}. Our previous work demonstrated that combining node transformer architectures with BERT (Bidirectional Encoder Representations from Transformers) sentiment analysis achieves 0.80\% mean absolute percentage error (MAPE) and 65\% directional accuracy (DA) on S\&P 500 stocks \cite{alridhawi2025nodeformer}. Yet this integrated model applies the same processing regardless of market conditions, leaving potential gains from regime-aware specialization unexploited.

The challenge of regime detection compounds prediction difficulties. Traditional approaches rely on hidden Markov models \cite{hamilton1989new} or threshold rules on volatility indicators \cite{liu2022vix}, both requiring manual specification of regime definitions. Supervised classifiers demand labeled training data identifying which historical periods constitute crises or anomalies. Such labels are subjective, backward-looking, and fail to generalize as market structure evolves. A system that automatically discovers regime boundaries from prediction performance itself would avoid these limitations.

This paper introduces an adaptive framework addressing both challenges. An autoencoder trained on normal market data learns to reconstruct typical price patterns; high reconstruction error indicates departure from learned normality. This weakly supervised anomaly score gates data flow through dual node transformer pathways: one optimized for stable conditions, another incorporating event-specific features for turbulent periods. A Soft Actor-Critic (SAC) reinforcement learning controller observes prediction outcomes and adjusts the autoencoder threshold and pathway blending weights to maximize forecasting accuracy. The SAC component adapts the anomaly-routing threshold by discovering which threshold settings improve downstream predictions.

The contributions of this work are:

\begin{enumerate}
\item An autoencoder-based regime detection mechanism that identifies market state shifts using weakly supervised anomaly detection trained on historically stable market periods. The autoencoder learns a compressed representation of normal market dynamics; deviations from this representation trigger event-aware processing.

\item A dual node transformer architecture with specialized pathways for stable and volatile market conditions. The event pathway incorporates additional features including volatility regime indicators, sentiment spikes, and event characterization signals.

\item A Soft Actor-Critic reinforcement learning controller that adaptively tunes the regime detection threshold and pathway blending based on realized prediction performance. This enables the system to learn adaptive regime definitions from prediction outcomes rather than relying on fully hand-labeled regime annotations.

\item Experimental validation demonstrating a 26\% MAPE reduction over the baseline integrated node transformer (0.59\% vs 0.80\%) and a 7 percentage point improvement in directional accuracy (72\% vs 65\%).
\end{enumerate}

Section~\ref{sec:literature} reviews related work on regime detection, adaptive prediction, and reinforcement learning for financial applications. Section~\ref{sec:methodology} presents the proposed architecture. Section~\ref{sec:experiments} reports experimental results, and Section~\ref{sec:discussion} discusses findings, limitations, and implications.

%========================================
% SECTION 2: LITERATURE REVIEW
%========================================
\section{Literature Review}
\label{sec:literature}

\subsection{Regime Detection in Financial Markets}

Market regime identification has a long history in econometrics and quantitative finance. Hamilton \cite{hamilton1989new} introduced Markov-switching models that probabilistically transition between states with distinct statistical properties. These models estimate regime-specific parameters (means, variances, transition probabilities) via maximum likelihood, enabling classification of historical periods into regimes. Extensions incorporate time-varying transition probabilities \cite{diebold1994regime} and multivariate dependencies.

Threshold models offer an alternative where regime switches occur when observable variables cross specified boundaries. The Self-Exciting Threshold Autoregressive (SETAR) model \cite{tong1990non} switches dynamics based on lagged values of the series itself. In finance, volatility indices such as VIX (CBOE Volatility Index) commonly serve as regime indicators, with thresholds separating low, medium, and high volatility states.

Machine learning approaches to regime detection include clustering methods that partition historical periods based on feature similarity \cite{nystrup2017clustering}, hidden Markov models with neural network emission distributions, and change-point detection algorithms \cite{aminikhanghahi2017survey}. These methods generally require either explicit labels or assumptions about the number and nature of regimes. The framework proposed in this paper does not entirely avoid such assumptions, as the primary routing is binary and the event pathway conditions on three VIX-based volatility levels. This design nonetheless requires fewer structural commitments than methods that must specify the number, boundaries, and statistical properties of multiple regime states. The autoencoder learns to distinguish normal from anomalous market conditions through reconstruction error without requiring explicit regime definitions, and the SAC controller continuously adapts the routing threshold based on prediction feedback rather than relying on fixed, manually chosen boundaries. The regime structure is therefore partially discovered from data rather than imposed entirely by the modeler.

\subsection{Autoencoders for Anomaly Detection}

Autoencoders learn compressed representations by reconstructing inputs through an information bottleneck \cite{hinton2006reducing}. The encoder maps inputs to a lower-dimensional latent space, and the decoder reconstructs the original input from this representation. When trained on normal data, autoencoders reconstruct typical patterns with low error; anomalous inputs that deviate from the training distribution yield higher reconstruction error, providing an anomaly score.

Variational autoencoders (VAEs) extend this framework by imposing distributional constraints on the latent space \cite{kingma2014auto}. The VAE objective combines reconstruction loss with a regularization term encouraging the latent distribution to match a prior (typically standard Gaussian). This probabilistic formulation enables principled uncertainty quantification and generation of novel samples.

In financial applications, autoencoders have been applied to fraud detection \cite{pumsirirat2018credit}, anomaly identification in trading patterns \cite{ahmed2016survey}, and feature extraction for downstream prediction tasks \cite{bao2017deep}. Liu et al. \cite{liu2022bilstm} employed autoencoder-based feature extraction combined with bidirectional LSTM (Long Short-Term Memory) for stock price prediction, reporting improved performance from the learned representations.

\subsection{Graph Neural Networks for Stock Prediction}

Graph neural networks (GNNs) model relational structure among entities through message passing over graph topology \cite{wu2020comprehensive}. In stock prediction, nodes represent individual securities while edges capture relationships including sectoral affiliation, supply chain connections, or return correlations.

Chen et al. \cite{chen2021graph} proposed a graph convolutional feature-based CNN combining graph convolutions with dual convolutional networks for market-level and stock-level features. Wang et al. \cite{wang2022multigraph} introduced multi-graph architectures defining both static (sector) and dynamic (correlation) graphs, achieving 5.11\% error reduction over LSTM baselines on Chinese market indices.

The node transformer architecture \cite{wu2022nodeformer} extends transformers to graph-structured data through attention mechanisms that respect graph topology. Unlike standard graph neural networks with fixed message-passing schemes, node transformers learn contextualized representations through adaptive attention over graph neighborhoods.

\subsection{Reinforcement Learning in Finance}

Reinforcement learning (RL) optimizes sequential decision-making through interaction with an environment, learning policies that maximize cumulative reward \cite{sutton2018reinforcement}. Financial applications include portfolio optimization \cite{jiang2017deep}, order execution \cite{ning2021double}, and trading strategy development \cite{deng2016deep}.

Deep RL algorithms combine neural network function approximation with RL principles. Deep Q-Networks (DQN) learn action-value functions for discrete action spaces \cite{mnih2015human}, while policy gradient methods directly optimize policies for continuous action spaces. Actor-critic algorithms unify both approaches by combining value estimation (critic) with policy optimization (actor) for improved stability and sample efficiency.

Soft Actor-Critic (SAC) \cite{haarnoja2018soft} incorporates entropy regularization into the actor-critic framework, encouraging exploration while maintaining policy stability. By adding policy entropy to the reward, the maximum entropy objective prevents premature convergence to deterministic policies, and SAC performs well across continuous control tasks with delayed, noisy reward signals \cite{haarnoja2018soft, haarnoja2019sac}.

Here, SAC serves not as a trading agent but as a meta-controller that learns to configure the prediction system itself. The controller adjusts the autoencoder threshold and pathway blending weights based on observed prediction performance, effectively learning what regime definitions optimize downstream forecasting accuracy.

\subsection{Research Positioning}

Prior work has addressed regime detection and stock prediction as separate problems. Regime-switching models identify market states but do not adapt prediction methods accordingly \cite{hamilton1989new, ang2002international, guidolin2007asset}, while stock prediction models treat all conditions uniformly or rely on hand-crafted regime indicators \cite{fischer2018deep, chen2021graph}. Our framework integrates these elements by pairing unsupervised regime detection (autoencoder) with specialized prediction pathways (dual node transformers) and a SAC controller that learns optimal regime definitions from prediction outcomes. This closed-loop architecture enables the system to discover useful regime boundaries rather than imposing them a priori.

%========================================
% SECTION 3: METHODOLOGY
%========================================
\section{Methodology}
\label{sec:methodology}

\subsection{System Overview}

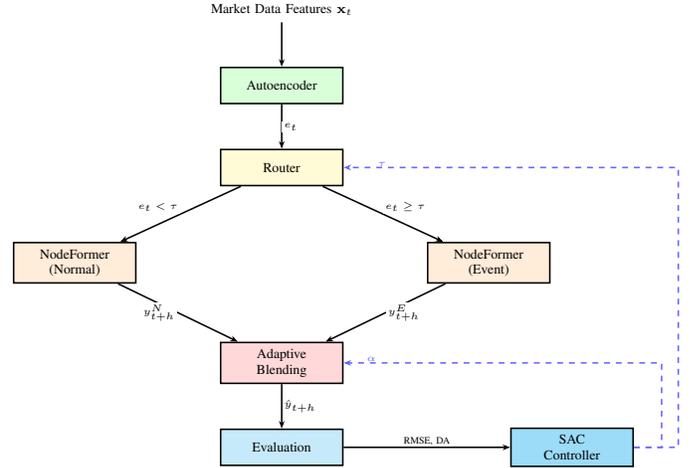
\begin{figure}[!htbp]
\centering
\resizebox{\columnwidth}{!}{%
\begin{tikzpicture}[
    node distance=0.6cm and 0.8cm,
    box/.style={rectangle, draw, minimum width=2.2cm, minimum height=0.65cm, align=center, font=\scriptsize, thick},
    greenbox/.style={box, fill=green!15},
    orangebox/.style={box, fill=orange!15},
    redbox/.style={box, fill=red!15},
    yellowbox/.style={box, fill=yellow!20},
    cyanbox/.style={box, fill=cyan!20},
    arrow/.style={-{Stealth[scale=0.6]}, thick},
    dashedarrow/.style={-{Stealth[scale=0.6]}, thick, dashed, blue!60},
    lbl/.style={font=\tiny, fill=white, inner sep=1pt}
]

% Input - plain text, no box (input data, not a component)
\node[font=\scriptsize] (input) {Market Data Features $\vect{x}_t$};

% Components
\node[greenbox, below=0.8cm of input] (ae) {Autoencoder};
\node[yellowbox, below=0.8cm of ae] (router) {Router};
\node[orangebox, below left=1.0cm and 1.5cm of router] (normal) {NodeFormer\\(Normal)};
\node[orangebox, below right=1.0cm and 1.5cm of router] (event) {NodeFormer\\(Event)};
\node[redbox, below=2.8cm of router] (fusion) {Adaptive\\Blending};
\node[cyanbox, below=0.8cm of fusion] (eval) {Evaluation};
\node[box, fill=cyan!35, right=3.0cm of eval] (sac) {SAC\\Controller};

% Main flow arrows with output labels on arrows
\draw[arrow] (input) -- (ae);
\draw[arrow] (ae) -- node[lbl, right] {$e_t$} (router);
\draw[arrow] (router) -- node[lbl, above left] {$e_t < \tau$} (normal);
\draw[arrow] (router) -- node[lbl, above right] {$e_t \geq \tau$} (event);
\draw[arrow] (normal) -- node[lbl, left] {$y^{N}_{t+h}$} (fusion);
\draw[arrow] (event) -- node[lbl, right] {$y^{E}_{t+h}$} (fusion);
\draw[arrow] (fusion) -- node[lbl, right] {$\hat{y}_{t+h}$} (eval);

% Arrow from eval to SAC with performance metrics
\draw[arrow] (eval.east) -- node[lbl, above] {RMSE, DA} (sac.west);

% Feedback arrows from SAC
\draw[dashedarrow] (sac.east) -- ++(0.8,0) |- (router.east);
\draw[dashedarrow] (sac.east) -- ++(0.5,0) |- (fusion.east);

% Labels for feedback
\node[font=\tiny, blue!60, above right=-0.1cm and 0.5cm of router.east] {$\tau$};
\node[font=\tiny, blue!60, above right=-0.1cm and 0.3cm of fusion.east] {$\alpha$};

\end{tikzpicture}%
}
\caption{System architecture overview. Market features $\vect{x}_t$ enter the autoencoder, which produces reconstruction error $e_t$ (shown on arrow). The router directs data to normal or event node transformer pathways based on whether $e_t$ exceeds the learned threshold $\tau$. Each pathway produces a prediction ($y^{N}_{t+h}$, $y^{E}_{t+h}$), and adaptive blending combines them into the final forecast $\hat{y}_{t+h}$. The SAC controller observes evaluation metrics and adjusts both $\tau$ and $\alpha$ (dashed blue arrows) to optimize forecasting accuracy.}
\label{fig:system_architecture}
\end{figure}

Figure~\ref{fig:system_architecture} presents the complete system architecture. Raw market data flows through feature engineering to produce technical indicators and normalized price features. The autoencoder processes these features, producing a reconstruction error score that quantifies deviation from normal market patterns and serves as the basis for regime classification. Based on this score and a learned threshold, the router directs data to either the normal or event node transformer pathway depending on the detected regime. Both pathways produce predictions that are blended according to learned weights. The final prediction is evaluated, and the SAC controller uses this feedback to adjust the autoencoder threshold and blending parameters for subsequent iterations.

The term \emph{regime} in this framework operates at two distinct levels. At the primary level, the autoencoder performs a binary classification of each trading day as either \emph{normal} or \emph{anomalous} based on whether its reconstruction error exceeds a learned threshold $\tau$. This binary decision determines routing: days classified as normal are processed by the normal node transformer, while anomalous days are directed to the event node transformer. A binary primary classification is chosen rather than a multi-class scheme for both practical and theoretical reasons. The autoencoder's reconstruction error is a scalar anomaly score that naturally lends itself to thresholding rather than clustering into multiple categories, and the fundamental distinction in anomaly detection is between in-distribution and out-of-distribution inputs. Attempting to subdivide anomalous states at the routing stage would require assumptions about the number and nature of anomaly categories that the unsupervised autoencoder is not designed to make; instead, that finer-grained characterization is deferred to the event pathway itself.

Within the event pathway, a secondary level of regime characterization captures the heterogeneity of anomalous periods. The event context vector $\vect{c}_t$ provides the event node transformer with descriptive features including a VIX-based volatility classification into three levels (low, medium, or high, determined by training-period terciles), sentiment spike indicators, earnings event proximity, and cross-asset stress measures. This secondary characterization does not constitute a separate routing mechanism; rather, it conditions the event transformer's internal representations by supplying information about the nature of the detected anomaly. An earnings-driven disruption during a period of otherwise low market volatility produces different price dynamics than a systemic sell-off during an already elevated volatility regime, and the context vector enables the transformer to learn these distinctions from the data. The normal transformer does not receive this additional context because it processes in-distribution samples where market dynamics follow the stable patterns learned during the autoencoder's training phase, making regime-specific conditioning unnecessary.

\subsection{Feature Engineering}

\begin{figure}[!htbp]
\centering
\resizebox{0.9\columnwidth}{!}{%
\begin{tikzpicture}[
    node distance=0.5cm and 0.8cm,
    box/.style={rectangle, draw, minimum width=1.8cm, minimum height=0.6cm, align=center, font=\tiny, thick},
    smallbox/.style={rectangle, draw, minimum width=1.4cm, minimum height=0.5cm, align=center, font=\tiny, thick},
    arrow/.style={-{Stealth[scale=0.5]}, thick}
]

% Raw data input - centered at top
\node[box, fill=blue!15] (raw) {Raw OHLCV\\Data};

% Technical indicators branch - spread horizontally with more spacing
\node[smallbox, fill=green!15, below=1.2cm of raw, xshift=-3.0cm] (sma) {SMA\\5, 10, 20};
\node[smallbox, fill=green!15, below=1.2cm of raw, xshift=-1.0cm] (ema) {EMA\\5, 10, 20};
\node[smallbox, fill=green!15, below=1.2cm of raw, xshift=1.0cm] (momentum) {RSI, MACD\\Returns};
\node[smallbox, fill=yellow!25, below=1.2cm of raw, xshift=3.0cm] (vol) {Rolling\\Volatility};

% Normalization - centered below indicators
\node[box, fill=orange!20, below=1.2cm of raw, yshift=-1.5cm, minimum width=2.5cm] (norm) {Z-Score Normalization\\(Expanding Window)};

% Output vectors - spread below normalization
\node[box, fill=red!15, below=0.8cm of norm, xshift=-2.0cm] (xpred) {Prediction Features\\$\vect{x}_{i,t} \in \mathbb{R}^{17}$};
\node[box, fill=red!15, below=0.8cm of norm, xshift=2.0cm] (xrouter) {Router Features\\$\vect{x}_{i,t}^{\text{router}} \in \mathbb{R}^{6}$};

% Arrows from raw to indicators - go down then spread out
\draw[arrow] (raw.south) -- ++(0,-0.4) -| (sma.north);
\draw[arrow] (raw.south) -- ++(0,-0.4) -| (ema.north);
\draw[arrow] (raw.south) -- ++(0,-0.4) -| (momentum.north);
\draw[arrow] (raw.south) -- ++(0,-0.4) -| (vol.north);

% Arrows from indicators to normalization
\draw[arrow] (sma.south) -- ++(0,-0.3) -| (norm.north west);
\draw[arrow] (ema.south) -- ++(0,-0.3) -| ([xshift=-0.3cm]norm.north);
\draw[arrow] (momentum.south) -- ++(0,-0.3) -| ([xshift=0.3cm]norm.north);
\draw[arrow] (vol.south) -- ++(0,-0.3) -| (norm.north east);

% Arrows to output
\draw[arrow] (norm.south) -- ++(0,-0.3) -| (xpred.north);
\draw[arrow] (norm.south) -- ++(0,-0.3) -| (xrouter.north);

\end{tikzpicture}%
}
\caption{Feature engineering pipeline. Raw OHLCV data is processed through technical indicator computations (SMA, EMA, RSI, MACD, volatility). All features undergo expanding-window z-score normalization to prevent look-ahead bias, producing prediction features and router-specific features.}
\label{fig:feature_pipeline}
\end{figure}
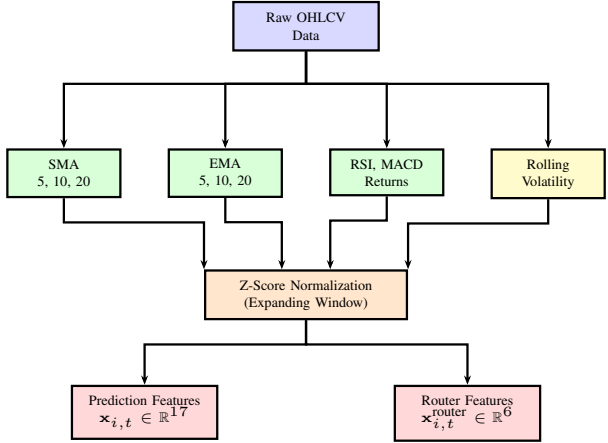

Input features follow established methodologies for financial time series. For each stock $i$ at time $t$, the raw feature vector comprises:

\begin{equation}
\vect{x}_{i,t}^{\text{raw}} = [O_t, H_t, L_t, C_t, V_t]
\end{equation}

where $O_t$, $H_t$, $L_t$, $C_t$ denote open, high, low, and closing prices, and $V_t$ is trading volume (collectively referred to as OHLCV data). Technical indicators include simple moving averages (SMA) at 5, 10, and 20-day windows, exponential moving averages (EMA) at matching windows, 14-day Relative Strength Index (RSI), Moving Average Convergence Divergence (MACD) with standard (12, 26, 9) parameters, daily returns, log returns, and 20-day rolling volatility. Figure~\ref{fig:feature_pipeline} illustrates this pipeline.

In addition to prediction features, router-specific features capture regime-relevant signals:

\begin{equation}
\vect{x}_{i,t}^{\text{router}} = [\sigma_t^{(5)}, \sigma_t^{(20)}, \Delta\text{VIX}_t, \rho_t^{\Delta}, |S_t|, \nu_t^{\text{post}}]
\end{equation}

where $\sigma_t^{(k)}$ is $k$-day rolling volatility, $\Delta\text{VIX}_t$ is VIX percentage change, $\rho_t^{\Delta}$ is change in average pairwise correlation among stocks, $|S_t|$ is absolute sentiment magnitude, and $\nu_t^{\text{post}}$ is post velocity (count of X posts mentioning crisis-related keywords within the trading day). Sentiment enters the router as an absolute value because the router's function is anomaly detection rather than directional prediction. For the purpose of identifying unusual market conditions, the magnitude of sentiment deviation is the relevant signal: both strongly negative sentiment (indicating panic) and strongly positive sentiment (indicating euphoria or speculative excess) represent departures from typical market behavior. The signed sentiment score $S_t$ is retained in the full prediction feature vector $\vect{x}_{i,t}$ that reaches the node transformers, so directional information contributes to the price forecasts themselves even though the routing decision depends only on sentiment intensity. These six router features are chosen to capture both gradual shifts (rolling volatility, correlation changes) and abrupt events (VIX spikes, sentiment surges, social media clustering).

Missing values in price data (due to trading halts or data gaps) are handled through temporally-aware imputation. For training data, short gaps (1-2 trading days) use linear interpolation between surrounding known values. For validation and test data, only forward-filling from the most recent observed value is applied to ensure no future information leaks into predictions. Technical indicators (SMA, EMA, RSI, MACD) are computed only after imputation, using the forward-filled values.

Normalization uses expanding-window z-scores to prevent look-ahead bias. During training, each feature is standardized using the mean and standard deviation computed over all available data from the start of the training period up to time $t$:

\begin{equation}
\tilde{x}_{i,t} = \frac{x_{i,t} - \mu_{1:t}}{\sigma_{1:t}}
\end{equation}

where $\mu_{1:t}$ and $\sigma_{1:t}$ are the cumulative mean and standard deviation from the first training observation through time $t$. This expanding window ensures that normalization at each time step uses only past information. During validation and testing, normalization statistics are fixed at the full training-period values ($\mu_{1:T_{\text{train}}}$ and $\sigma_{1:T_{\text{train}}}$), ensuring that no information from the evaluation period influences standardization.

\subsection{Autoencoder for Regime Detection}
\label{subsec:autoencoder}

\begin{figure}[!htbp]
\centering
\resizebox{\columnwidth}{!}{%
\begin{tikzpicture}[
    node distance=0.5cm,
    layer/.style={rectangle, draw, minimum width=0.8cm, minimum height=1.8cm, align=center, font=\tiny, thick},
    smalllayer/.style={rectangle, draw, minimum width=0.6cm, minimum height=1.2cm, align=center, font=\tiny, thick},
    tinylayer/.style={rectangle, draw, minimum width=0.5cm, minimum height=0.8cm, align=center, font=\tiny, thick},
    arrow/.style={-{Stealth[scale=0.5]}, thick},
    label/.style={font=\tiny}
]

% Input
\node[layer, fill=blue!15, minimum height=2.4cm] (input) {$\vect{x}_t$\\[0.1cm]$d_{\text{in}}$};

% Encoder layers
\node[layer, fill=green!20, right=0.6cm of input, minimum height=2.0cm] (enc1) {64};
\node[smalllayer, fill=green!25, right=0.5cm of enc1] (enc2) {32};

% Latent
\node[tinylayer, fill=yellow!30, right=0.6cm of enc2] (latent) {$\vect{z}_t$\\$d_z$};

% Decoder layers
\node[smalllayer, fill=orange!20, right=0.6cm of latent] (dec1) {32};
\node[layer, fill=orange!25, right=0.5cm of dec1, minimum height=2.0cm] (dec2) {64};

% Output
\node[layer, fill=red!15, right=0.6cm of dec2, minimum height=2.4cm] (output) {$\hat{\vect{x}}_t$\\[0.1cm]$d_{\text{in}}$};

% Reconstruction error
\node[rectangle, draw, fill=purple!15, right=0.8cm of output, minimum width=1.2cm, minimum height=0.8cm, align=center, font=\tiny, thick] (error) {$e_t = \|\vect{x}_t - \hat{\vect{x}}_t\|_2^2$};

% Arrows
\draw[arrow] (input) -- node[above, label] {$\mat{W}_1$} (enc1);
\draw[arrow] (enc1) -- node[above, label] {$\mat{W}_2$} (enc2);
\draw[arrow] (enc2) -- (latent);
\draw[arrow] (latent) -- (dec1);
\draw[arrow] (dec1) -- node[above, label] {$\mat{W}_3$} (dec2);
\draw[arrow] (dec2) -- node[above, label] {$\mat{W}_4$} (output);
\draw[arrow] (output) -- (error);

% Labels
\node[below=0.3cm of enc1, label] {ReLU};
\node[below=0.3cm of enc2, label] {ReLU};
\node[below=0.3cm of dec1, label] {ReLU};
\node[below=0.3cm of dec2, label] {ReLU};

% Brace for encoder
\draw[decorate, decoration={brace, amplitude=5pt, raise=2pt}] (input.north west) -- (enc2.north east) node[midway, above=8pt, font=\scriptsize] {Encoder $f_{\text{enc}}$};

% Brace for decoder
\draw[decorate, decoration={brace, amplitude=5pt, raise=2pt}] (dec1.north west) -- (output.north east) node[midway, above=8pt, font=\scriptsize] {Decoder $f_{\text{dec}}$};

\end{tikzpicture}%
}
\caption{Autoencoder architecture for regime detection. The encoder compresses the input feature vector through two hidden layers (64, 32 units) to a latent representation $\vect{z}_t$ of dimension $d_z = 32$. The decoder reconstructs the input through symmetric layers. Reconstruction error $e_t$ serves as the anomaly score for regime classification.}
\label{fig:autoencoder_arch}
\end{figure}
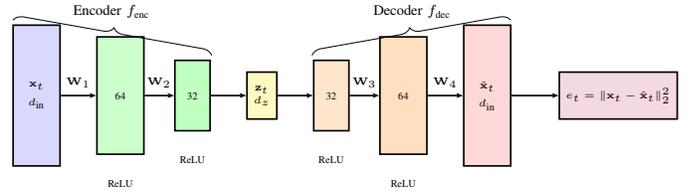

The autoencoder learns a compressed representation of normal market dynamics. It is trained exclusively on data from stable market periods, defined during training as days where VIX remains below the 75th percentile of its training-period distribution. Figure~\ref{fig:autoencoder_arch} presents the detailed architecture.

\subsubsection{Architecture and Training}

The encoder maps the concatenated feature vector to a latent representation through two hidden layers:

\begin{equation}
\vect{z}_t = f_{\text{enc}}(\vect{x}_t) = \text{ReLU}(\mat{W}_2 \cdot \text{ReLU}(\mat{W}_1 \vect{x}_t + \vect{b}_1) + \vect{b}_2)
\end{equation}

where $\vect{x}_t \in \mathbb{R}^{d_{\text{in}}}$ is the input feature vector, $\vect{z}_t \in \mathbb{R}^{d_z}$ is the latent representation with $d_z = 32$, and $\mat{W}_1 \in \mathbb{R}^{64 \times d_{\text{in}}}$, $\mat{W}_2 \in \mathbb{R}^{32 \times 64}$ are weight matrices. The decoder reconstructs the input through a symmetric architecture:

\begin{equation}
\hat{\vect{x}}_t = f_{\text{dec}}(\vect{z}_t) = \mat{W}_4 \cdot \text{ReLU}(\mat{W}_3 \vect{z}_t + \vect{b}_3) + \vect{b}_4
\end{equation}

where $\mat{W}_3 \in \mathbb{R}^{64 \times 32}$ and $\mat{W}_4 \in \mathbb{R}^{d_{\text{in}} \times 64}$. The autoencoder is trained to minimize reconstruction loss over the stable-period data:

\begin{equation}
\mathcal{L}_{\text{AE}} = \frac{1}{T} \sum_{t=1}^{T} \|\vect{x}_t - \hat{\vect{x}}_t\|_2^2
\end{equation}

Training uses the Adam optimizer with learning rate $10^{-3}$, batch size 64, for a maximum of 20 epochs with early stopping based on validation reconstruction loss.

\subsubsection{Anomaly Score and Routing}

At inference time, the reconstruction error serves as an anomaly score $e_t = \|\vect{x}_t - \hat{\vect{x}}_t\|_2^2$. Data points with $e_t$ exceeding threshold $\tau$ are classified as anomalous and routed to the event pathway, while those below $\tau$ proceed through the normal pathway. The threshold $\tau$ is initialized at the 95th percentile of training-set reconstruction errors and subsequently adjusted by the SAC controller.

\subsection{Dual Node Transformer Architecture}

\begin{figure}[!htbp]
\centering
\resizebox{\columnwidth}{!}{%
\begin{tikzpicture}[
    node distance=0.4cm and 0.5cm,
    box/.style={rectangle, draw, minimum width=2.0cm, minimum height=0.5cm, align=center, font=\tiny, thick},
    wideboxL/.style={rectangle, draw, minimum width=2.8cm, minimum height=0.5cm, align=center, font=\tiny, thick, fill=orange!15},
    wideboxR/.style={rectangle, draw, minimum width=2.8cm, minimum height=0.5cm, align=center, font=\tiny, thick, fill=blue!15},
    arrow/.style={-{Stealth[scale=0.5]}, thick}
]

% Router at top
\node[box, fill=yellow!25] (router) {Router: $e_t \gtrless \tau$};

% Left pathway label
\node[below left=0.6cm and 1.8cm of router, font=\scriptsize\bfseries] (leftlabel) {Normal Path ($e_t < \tau$)};

% Right pathway label
\node[below right=0.6cm and 1.8cm of router, font=\scriptsize\bfseries] (rightlabel) {Event Path ($e_t \geq \tau$)};

% Left pathway components
\node[wideboxL, below=0.3cm of leftlabel] (linput) {Input: $\vect{x}_{i,t}$};
\node[wideboxL, below=0.3cm of linput] (lembed) {Stock Embedding};
\node[wideboxL, below=0.3cm of lembed] (ltemporal) {Temporal Encoding};
\node[wideboxL, below=0.3cm of ltemporal] (lmha1) {Multi-Head Attention $\times$ 6};
\node[wideboxL, below=0.3cm of lmha1] (lffn) {Feed-Forward + LayerNorm};
\node[wideboxL, below=0.3cm of lffn] (lpred) {Prediction Head};
\node[wideboxL, below=0.3cm of lpred] (lout) {$y^{\text{normal}}_{i,t+h}$};

% Right pathway components
\node[wideboxR, below=0.3cm of rightlabel] (rinput) {Input: $[\vect{x}_{i,t} \| \vect{c}_t]$};
\node[wideboxR, below=0.3cm of rinput] (rembed) {Stock + Event Embedding};
\node[wideboxR, below=0.3cm of rembed] (rtemporal) {Temporal Encoding};
\node[wideboxR, below=0.3cm of rtemporal] (rmha1) {Multi-Head Attention $\times$ 6};
\node[wideboxR, below=0.3cm of rmha1] (rffn) {Feed-Forward + LayerNorm};
\node[wideboxR, below=0.3cm of rffn] (rpred) {Prediction Head};
\node[wideboxR, below=0.3cm of rpred] (rout) {$y^{\text{event}}_{i,t+h}$};

% Router arrows
\draw[arrow] (router.south) -- ++(-2.5,-0.3) -- (leftlabel.north);
\draw[arrow] (router.south) -- ++(2.5,-0.3) -- (rightlabel.north);

% Left pathway arrows
\draw[arrow] (linput) -- (lembed);
\draw[arrow] (lembed) -- (ltemporal);
\draw[arrow] (ltemporal) -- (lmha1);
\draw[arrow] (lmha1) -- (lffn);
\draw[arrow] (lffn) -- (lpred);
\draw[arrow] (lpred) -- (lout);

% Right pathway arrows
\draw[arrow] (rinput) -- (rembed);
\draw[arrow] (rembed) -- (rtemporal);
\draw[arrow] (rtemporal) -- (rmha1);
\draw[arrow] (rmha1) -- (rffn);
\draw[arrow] (rffn) -- (rpred);
\draw[arrow] (rpred) -- (rout);

% Fusion at bottom
\node[box, fill=red!20, below=1.0cm of router, yshift=-7.5cm] (fusion) {Adaptive Blending: $\hat{y} = \alpha y^{\text{normal}} + (1-\alpha) y^{\text{event}}$};

% Arrows to fusion
\draw[arrow] (lout.south) |- (fusion.west);
\draw[arrow] (rout.south) |- (fusion.east);

% Event context features box
\node[rectangle, draw, dashed, fill=gray!10, right=0.5cm of rinput, minimum width=2.2cm, minimum height=1.8cm, align=left, font=\tiny] (context) {
\textbf{Event Context $\vect{c}_t$:}\\
- Regime embedding\\
- Sentiment spike\\
- Days to earnings\\
- Cross-asset stress
};

\draw[arrow, dashed] (context.west) -- (rinput.east);

\end{tikzpicture}%
}
\caption{Dual node transformer architecture. The router directs data based on reconstruction error. The normal pathway (left, orange) processes typical market conditions with base features. The event pathway (right, blue) augments inputs with event context features $\vect{c}_t$. Both pathways follow the same architectural design (layer count, attention heads, model dimension) but maintain independently trained weights and differ in input dimensionality, as the event pathway accepts additional context features. Outputs are blended with adaptive weight $\alpha$.}
\label{fig:dual_nodeformer}
\end{figure}
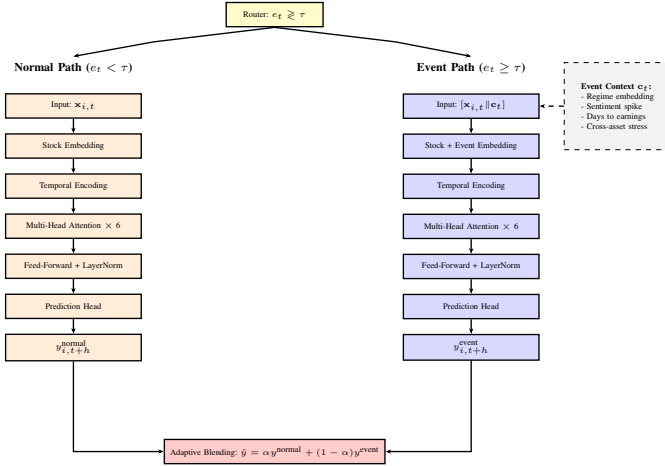

Two node transformer networks process data depending on regime classification. Both follow the same base architectural design (6 layers, 8 attention heads, 512 model dimension) but maintain independent weights trained on different data subsets, and the event pathway accepts a larger input due to additional context features. Figure~\ref{fig:dual_nodeformer} illustrates the dual pathway structure.

\subsubsection{Normal Node Transformer}

The normal pathway processes typical market conditions using the node transformer architecture \cite{wu2022nodeformer}, which extends standard transformers to graph-structured data by incorporating relational inductive biases into the attention mechanism. The stock market is represented as a graph $\mathcal{G} = (\mathcal{V}, \mathcal{E})$ with $N = 20$ stock nodes and a fully-connected edge set. While graph neural networks are often applied to larger graphs, the $N=20$ design balances cross-sectional breadth against temporal depth (252-day sequences per stock), and ablation results confirm that the graph structure contributes 7\% MAPE improvement (Table~\ref{tab:ablation}), indicating that cross-sectional dependencies carry predictive value even at this scale.

Each stock $i$ receives a learned embedding $\vect{s}_i \in \mathbb{R}^{d_s}$ that captures persistent stock-specific characteristics such as sector behavior and volatility profile. The input representation for stock $i$ at time $t$ concatenates the normalized feature vector with temporal encoding and the stock embedding:

\begin{equation}
\vect{h}_{i,t}^{(0)} = [\vect{x}_{i,t} \| \text{TE}(t) \| \vect{s}_i] \in \mathbb{R}^{d_{\text{in}}}
\end{equation}

Temporal encoding follows Vaswani et al. \cite{vaswani2017attention}, using sinusoidal positional encodings where $\text{TE}(t, 2k) = \sin(t / 10000^{2k/d})$ and $\text{TE}(t, 2k+1) = \cos(t / 10000^{2k/d})$ for dimension index $k$ and model dimension $d = 512$. This encoding allows the model to distinguish trading days and capture periodic patterns at multiple frequencies.

\begin{figure}[!htbp]
\centering
\resizebox{\columnwidth}{!}{%
\begin{tikzpicture}[
    node distance=1.0cm,
    stock/.style={circle, draw, minimum size=0.7cm, font=\tiny, thick},
    tech/.style={stock, fill=blue!25},
    fin/.style={stock, fill=green!25},
    health/.style={stock, fill=red!20},
    energy/.style={stock, fill=orange!30},
    consumer/.style={stock, fill=purple!20},
    edge/.style={-, thick},
    strongedge/.style={-, very thick}
]

% Technology cluster - spread out more
\node[tech] (aapl) at (0,0) {AAPL};
\node[tech] (msft) at (2.0,0.4) {MSFT};
\node[tech] (crm) at (1.0,1.8) {CRM};

% Financial - moved further left
\node[fin] (jpm) at (-1.8,-1.2) {JPM};

% Healthcare cluster - spread out more and moved right
\node[health] (jnj) at (4.5,0) {JNJ};
\node[health] (unh) at (6.0,0.8) {UNH};
\node[health] (pfe) at (5.2,-1.2) {PFE};

% Energy cluster - spread out more
\node[energy] (xom) at (-0.8,-2.8) {XOM};
\node[energy] (cvx) at (1.2,-3.0) {CVX};

% Consumer - spread out more
\node[consumer] (ko) at (3.2,-2.2) {KO};
\node[consumer] (pg) at (5.5,-1.8) {PG};

% Same-sector edges (strong)
\draw[strongedge, blue!60] (aapl) -- (msft) node[midway, above, font=\tiny] {0.78};
\draw[strongedge, blue!50] (aapl) -- (crm);
\draw[strongedge, blue!50] (msft) -- (crm);
\draw[strongedge, red!50] (jnj) -- (unh) node[midway, above, font=\tiny] {0.65};
\draw[strongedge, red!40] (jnj) -- (pfe);
\draw[strongedge, red!40] (unh) -- (pfe);
\draw[strongedge, orange!70] (xom) -- (cvx) node[midway, below, font=\tiny] {0.82};
\draw[strongedge, purple!50] (ko) -- (pg);

% Cross-sector edges (weak, dashed)
\draw[edge, gray!50, dashed] (aapl) -- (jpm);
\draw[edge, gray!40, dashed] (msft) -- (jnj);
\draw[edge, gray!40, dashed] (jpm) -- (xom);
\draw[edge, gray!30, dashed] (cvx) -- (ko);
\draw[edge, gray!30, dashed] (pfe) -- (pg);

% Legend - repositioned for new layout
\node[below=0.8cm of cvx, xshift=2.0cm, font=\scriptsize, align=left] (legend) {
\textcolor{blue!60}{$\blacksquare$} Technology \quad
\textcolor{green!60}{$\blacksquare$} Financial \quad
\textcolor{red!50}{$\blacksquare$} Healthcare \quad
\textcolor{orange!70}{$\blacksquare$} Energy \quad
\textcolor{purple!50}{$\blacksquare$} Consumer
};

\end{tikzpicture}%
}
\caption{Graph representation of stock relationships (representative subset of 11 stocks shown for clarity; full graph contains all 20 stocks). Nodes represent individual stocks, colored by sector. Solid edges indicate same-sector connections with higher learned weights (annotated values show correlation-based initialization from training data). Dashed edges represent weaker cross-sector correlations that are learned during training.}
\label{fig:graph_structure}
\end{figure}
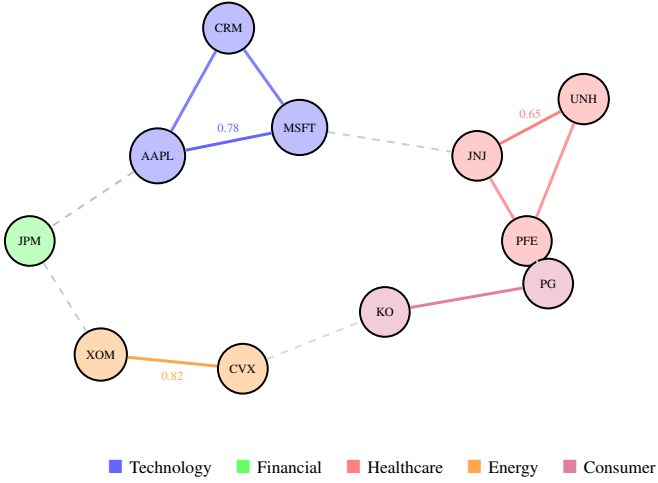

Edge weights in the graph are initialized from sector relationships and return correlations computed strictly on training data (1982-2010):

\begin{equation}
e_{ij}^{(0)} = 0.5 \cdot \delta_{\text{sector}}(i,j) + 0.5 \cdot \max(0, \rho_{ij}^{\text{train}})
\end{equation}

where $\delta_{\text{sector}}(i,j) = 1$ if stocks $i$ and $j$ share the same sector classification and $0$ otherwise, and $\rho_{ij}^{\text{train}}$ is the Pearson correlation of daily returns computed over the training period only, preventing any leakage from validation or test data. During training, edge weights are refined through a learnable function $e_{ij}^{(\ell+1)} = \sigma(\vect{w}_e^T [\vect{h}_i^{(\ell)} \| \vect{h}_j^{(\ell)}] + b_e)$, where $\sigma$ is the sigmoid function and $\vect{h}_i^{(\ell)}$ is the node representation at layer $\ell$. This allows the model to discover relationship patterns not captured by initial sector and correlation priors. Figure~\ref{fig:graph_structure} illustrates the resulting graph structure with representative edge weights.

At each layer, the node transformer applies multi-head self-attention with causal masking to jointly process all stocks across the temporal dimension. The input representations are projected into queries, keys, and values through learned linear transformations $\mat{Q} = \mat{X}\mat{W}^Q$, $\mat{K} = \mat{X}\mat{W}^K$, $\mat{V} = \mat{X}\mat{W}^V$, and attention output is computed as:

\begin{equation}
\mat{A} = \text{softmax}\left(\frac{\mat{Q}\mat{K}^T}{\sqrt{d_k}} + \mat{M} + \mat{E}\right)\mat{V}
\end{equation}

where $d_k = 64$ is the key dimension, $\mat{M}$ is the causal mask with $M_{ab} = -\infty$ if $a < b$ and $M_{ab} = 0$ otherwise, and $\mat{E} \in \mathbb{R}^{N \times N}$ is the learned edge weight matrix. The additive graph bias allows content-based attention (via $\mat{Q}\mat{K}^T$) and structural priors (via $\mat{E}$) to jointly determine how information flows between stocks at each layer, ensuring that predictions at time $t$ use only information from times up to and including $t$. The architecture uses $H = 8$ attention heads, each operating in 64 dimensions, yielding a total model dimension of $d_{\text{model}} = 512$.

\begin{figure}[!htbp]
\centering
\begin{tikzpicture}[
    scale=0.85,
    node distance=0.6cm,
    box/.style={rectangle, draw, minimum width=2.2cm, minimum height=0.5cm, align=center, font=\tiny, thick},
    smallbox/.style={rectangle, draw, minimum width=1.2cm, minimum height=0.4cm, align=center, font=\tiny, thick},
    arrow/.style={-{Stealth[scale=0.5]}, thick}
]

% Simple vertical layout - all centered
% Input at bottom
\node[box, fill=blue!10] (input) {Input $\vect{X}^{(\ell)}$};

% Multi-head attention block
\node[box, fill=orange!20, above=0.8cm of input] (mha) {Multi-Head Attention (8 heads)};

% Add & Norm 1
\node[smallbox, fill=gray!15, above=0.6cm of mha] (ln1) {Add \& LayerNorm};

% FFN
\node[box, fill=green!15, above=0.6cm of ln1] (ffn) {FFN (512$\rightarrow$2048$\rightarrow$512)};

% Add & Norm 2
\node[smallbox, fill=gray!15, above=0.6cm of ffn] (ln2) {Add \& LayerNorm};

% Output
\node[box, fill=red!10, above=0.8cm of ln2] (output) {Output $\vect{X}^{(\ell+1)}$};

% Main flow arrows (straight vertical)
\draw[arrow] (input) -- (mha);
\draw[arrow] (mha) -- (ln1);
\draw[arrow] (ln1) -- (ffn);
\draw[arrow] (ffn) -- (ln2);
\draw[arrow] (ln2) -- (output);

% Residual connection 1: input to ln1 - curved on the LEFT side
\draw[arrow, rounded corners=5pt] (input.west) -- ++(-1.5,0) |- (ln1.west);

% Residual connection 2: ln1 to ln2 - curved on the right side (shorter path)
\draw[arrow, rounded corners=5pt] (ln1.east) -- ++(0.8,0) |- (ln2.east);

% Annotations
\node[left=1.2cm of mha, font=\tiny, gray] {Dropout 0.1};
\node[right=0.5cm of ffn, font=\tiny, gray] {Dropout 0.1};
\node[right=0.3cm of output, font=\tiny] {$\times 6$ layers};

\end{tikzpicture}
\caption{Single transformer layer architecture. Input passes through multi-head self-attention, residual connection with layer normalization, feed-forward network, and another residual connection with normalization. The architecture stacks 6 such layers.}
\label{fig:transformer_layer}
\end{figure}
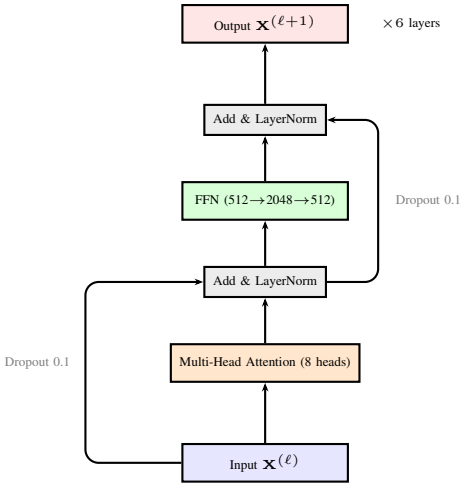

Each transformer layer follows the standard pre-norm residual pattern. The multi-head attention output is added to the input through a residual connection, followed by layer normalization. The normalized output then passes through a position-wise feed-forward network consisting of two linear transformations with a ReLU activation, expanding the representation to $d_{\text{ff}} = 2048$ dimensions before projecting back to 512. A second residual connection and layer normalization follow the feed-forward block. Dropout at rate 0.1 is applied after both the attention and feed-forward sublayers. The architecture stacks 6 such layers, with the output of the final layer fed into a prediction head consisting of a linear projection from the model dimension to a single scalar price prediction per stock. Figure~\ref{fig:transformer_layer} illustrates the detailed layer structure.

\subsubsection{Event Node Transformer}

The event pathway augments the base architecture with additional inputs capturing regime-specific information. The input to the event transformer concatenates the standard feature vector with an event context vector, $\vect{x}_{i,t}^{\text{event}} = [\vect{x}_{i,t} \| \vect{c}_t]$, where $\vect{c}_t \in \mathbb{R}^{d_c}$ with $d_c = 12$. This vector comprises four groups of features. A learned regime embedding $\vect{r}_t \in \mathbb{R}^{4}$ maps the current VIX regime (low, medium, or high, determined by training-period VIX terciles) through a trainable embedding layer. A sentiment spike component $\vect{s}_t \in \mathbb{R}^{2}$ encodes a binary flag and scaled magnitude when daily sentiment exceeds two standard deviations of training-period sentiment. An event characterization component $\vect{a}_t \in \mathbb{R}^{4}$ captures proximity to scheduled earnings announcements (days-to-announcement, normalized), historical earnings surprise magnitude for the stock, a binary earnings-window indicator, and sector-average surprise. Finally, a cross-asset stress vector $\bar{e}_t^{\text{cross}} \in \mathbb{R}^{2}$ encodes the mean and standard deviation of reconstruction error across all 20 stocks at time $t$, distinguishing systemic anomalies (high mean error) from idiosyncratic ones (high variance). The full context vector is the concatenation $\vect{c}_t = [\vect{r}_t \| \vect{s}_t \| \vect{a}_t \| \bar{e}_t^{\text{cross}}]$.

Architecturally, the event transformer differs from the normal pathway in two respects beyond its independently trained weights. First, the input projection layer is wider: while the normal transformer's first linear layer maps from $d_{\text{in}}$ dimensions (the concatenation of market features, temporal encoding, and stock embedding), the event transformer maps from $d_{\text{in}} + d_c$ dimensions to accommodate the appended context vector. This wider projection maps back to the shared model dimension of $d_{\text{model}} = 512$ before entering the first transformer layer, so all subsequent layers (the 6 transformer blocks, feed-forward networks, and prediction head) operate at the same dimensionality as the normal pathway. Second, the event pathway includes a trainable embedding layer that maps the discrete VIX regime label (one of three categories) to the continuous regime embedding $\vect{r}_t \in \mathbb{R}^4$. This embedding layer is an additional learnable component with no counterpart in the normal pathway, adding $3 \times 4 = 12$ trainable parameters for the three regime categories. The remaining context features ($\vect{s}_t$, $\vect{a}_t$, $\bar{e}_t^{\text{cross}}$) are continuous values that enter the context vector directly without additional learned transformations.

\subsubsection{Pathway Blending}

Rather than hard routing, predictions from both pathways are blended with an adaptive weight:

\begin{equation}
\hat{y}_{i,t+h} = \alpha_t \cdot y_{i,t+h}^{\text{normal}} + (1 - \alpha_t) \cdot y_{i,t+h}^{\text{event}}
\end{equation}

The blending coefficient $\alpha_t \in [0, 1]$ is determined by the SAC controller based on current market state. During high-confidence normal periods, $\alpha_t$ approaches 1; during clear anomalies, it approaches 0. Intermediate values enable smooth transitions and hedge against misclassification.

\subsection{Soft Actor-Critic Controller}
\label{subsec:sac}

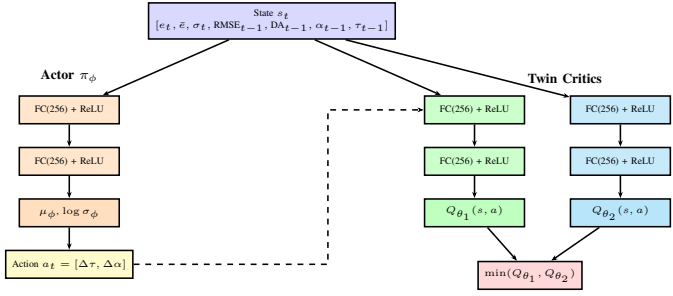
\begin{figure}[!htbp]
\centering
\resizebox{\columnwidth}{!}{%
\begin{tikzpicture}[
    node distance=0.5cm and 0.8cm,
    box/.style={rectangle, draw, minimum width=1.8cm, minimum height=0.5cm, align=center, font=\tiny, thick},
    widebox/.style={rectangle, draw, minimum width=2.4cm, minimum height=0.5cm, align=center, font=\tiny, thick},
    arrow/.style={-{Stealth[scale=0.5]}, thick}
]

% State input
\node[widebox, fill=blue!15] (state) {State $s_t$\\$[e_t, \bar{e}, \sigma_t, \text{RMSE}_{t-1}, \text{DA}_{t-1}, \alpha_{t-1}, \tau_{t-1}]$};

% Actor network
\node[box, fill=orange!20, below left=1.0cm and 0.5cm of state] (actor_h1) {FC(256) + ReLU};
\node[box, fill=orange!20, below=0.4cm of actor_h1] (actor_h2) {FC(256) + ReLU};
\node[box, fill=orange!25, below=0.4cm of actor_h2] (actor_out) {$\mu_\phi$, $\log\sigma_\phi$};
\node[box, fill=yellow!25, below=0.4cm of actor_out] (action) {Action $a_t = [\Delta\tau, \Delta\alpha]$};

% Critic networks (two)
\node[box, fill=green!20, below right=1.0cm and 0.5cm of state] (critic1_h1) {FC(256) + ReLU};
\node[box, fill=green!20, below=0.4cm of critic1_h1] (critic1_h2) {FC(256) + ReLU};
\node[box, fill=green!25, below=0.4cm of critic1_h2] (critic1_out) {$Q_{\theta_1}(s, a)$};

\node[box, fill=cyan!20, right=0.8cm of critic1_h1] (critic2_h1) {FC(256) + ReLU};
\node[box, fill=cyan!20, below=0.4cm of critic2_h1] (critic2_h2) {FC(256) + ReLU};
\node[box, fill=cyan!25, below=0.4cm of critic2_h2] (critic2_out) {$Q_{\theta_2}(s, a)$};

% Min operator
\node[box, fill=red!15, below=0.6cm of critic1_out, xshift=1.0cm] (minq) {$\min(Q_{\theta_1}, Q_{\theta_2})$};

% Labels
\node[above=0.1cm of actor_h1, font=\scriptsize\bfseries] {Actor $\pi_\phi$};
\node[above=0.1cm of critic2_h1, font=\scriptsize\bfseries, xshift=-1.0cm] {Twin Critics};

% Arrows from state
\draw[arrow] (state) -- (actor_h1);
\draw[arrow] (state) -- (critic1_h1);
\draw[arrow] (state) -- (critic2_h1);

\draw[arrow] (actor_h1) -- (actor_h2);
\draw[arrow] (actor_h2) -- (actor_out);
\draw[arrow] (actor_out) -- (action);

\draw[arrow] (critic1_h1) -- (critic1_h2);
\draw[arrow] (critic1_h2) -- (critic1_out);
\draw[arrow] (critic2_h1) -- (critic2_h2);
\draw[arrow] (critic2_h2) -- (critic2_out);

\draw[arrow] (critic1_out) -- (minq);
\draw[arrow] (critic2_out) -- (minq);

% Action to critics
\draw[arrow, dashed] (action.east) -- ++(3.5,0) |- (critic1_h1.west);

\end{tikzpicture}%
}
\caption{Soft Actor-Critic network architecture. The actor network (left, orange) maps states to a Gaussian policy over actions $[\Delta\tau, \Delta\alpha]$. Twin critic networks (right, green/cyan) estimate Q-values; the minimum is used to prevent overestimation. All networks use two hidden layers with 256 units and ReLU activation.}
\label{fig:sac_architecture}
\end{figure}

The SAC controller learns to configure the prediction system by adjusting the autoencoder threshold $\tau$ and blending weight $\alpha$ based on observed prediction performance. Although these are only two scalar parameters, the optimization landscape is non-trivial: the reward signal is delayed (prediction errors are observed only after the threshold decision), noisy (financial returns are inherently stochastic), and non-stationary (optimal thresholds shift as market regimes evolve). SAC is well suited to this setting because its entropy regularization prevents premature convergence to fixed threshold values, and its off-policy learning with experience replay enables sample-efficient adaptation from sparse, delayed feedback. Simpler alternatives such as grid search or bandit methods typically assume stationary reward distributions \cite{sutton2018reinforcement} and cannot adapt continuously to shifting regime dynamics. Figure~\ref{fig:sac_architecture} presents the actor-critic network architecture.

\subsubsection{Markov Decision Process Formulation}

The control problem is formulated as a Markov Decision Process (MDP). The state $s_t$ comprises:

\begin{equation}
s_t = [e_t, \bar{e}_{t-k:t}, \sigma_t, \text{RMSE}_{t-1}, \text{DA}_{t-1}, \alpha_{t-1}, \tau_{t-1}]
\end{equation}

including current reconstruction error, recent error history over the past $k = 5$ trading days (one week), volatility, previous prediction metrics (Root Mean Squared Error (RMSE) and Directional Accuracy (DA)), and current parameter settings. The action space consists of continuous adjustments $a_t = [\Delta\tau, \Delta\alpha] \in [-0.1, 0.1]^2$ to threshold and blending weight, clipped to maintain $\tau \in [e_{\text{min}}, e_{\text{max}}]$ and $\alpha \in [0, 1]$. The reward signal combines prediction accuracy and stability:

\begin{equation}
r_t = -\text{RMSE}_t - \lambda_{\text{dir}} \cdot (1 - \text{DA}_t) - \lambda_{\text{stable}} \cdot |\Delta\tau|
\end{equation}

where $\lambda_{\text{dir}} = 0.5$ weights directional accuracy and $\lambda_{\text{stable}} = 0.1$ penalizes threshold instability to prevent oscillation.

\subsubsection{SAC Algorithm}

SAC maximizes the entropy-regularized objective:

\begin{equation}
J(\pi) = \sum_{t=0}^{T} \mathbb{E}\left[r_t + \alpha_{\text{ent}} \mathcal{H}(\pi(\cdot|s_t))\right]
\end{equation}

where $\mathcal{H}$ is policy entropy and $\alpha_{\text{ent}}$ is the temperature parameter controlling exploration. The actor network $\pi_\phi(a|s)$ outputs a Gaussian distribution over actions, $\pi_\phi(a|s) = \mathcal{N}(\mu_\phi(s), \sigma_\phi(s)^2)$, while two critic networks $Q_{\theta_1}$, $Q_{\theta_2}$ estimate action values. To prevent overestimation, the minimum of both critics is used:

\begin{equation}
Q(s, a) = \min(Q_{\theta_1}(s, a), Q_{\theta_2}(s, a))
\end{equation}

All networks are feed-forward with two hidden layers of 256 units each. Training uses the Adam optimizer with learning rate $3 \times 10^{-4}$, soft target updates with $\tau_{\text{soft}} = 0.005$, and replay buffer size $10^5$.

\subsubsection{Training Protocol}

The SAC controller is trained after the autoencoder and node transformers are pre-trained. Training begins by initializing $\tau$ at the 95th percentile of training reconstruction errors and $\alpha = 0.5$. At each step, the controller computes predictions using the current $\tau$ and $\alpha$, evaluates them against actual outcomes to obtain the reward signal, updates the SAC networks from collected transitions, and applies the learned action adjustments to both parameters. This loop runs for 50 epochs with 1000 steps per epoch. Temperature $\alpha_{\text{ent}}$ is automatically tuned to target entropy $-\dim(a)$ following Haarnoja et al. \cite{haarnoja2018soft}.

\subsection{Training Pipeline}

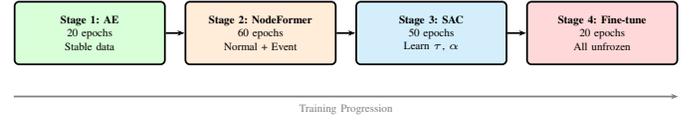
\begin{figure}[!htbp]
\centering
\resizebox{\columnwidth}{!}{%
\begin{tikzpicture}[
    node distance=0.3cm and 0.2cm,
    stage/.style={rectangle, draw, minimum width=2.4cm, minimum height=1.0cm, align=center, font=\tiny, thick, rounded corners=2pt},
    arrow/.style={-{Stealth[scale=0.5]}, thick}
]

% Stage 1
\node[stage, fill=green!15] (s1) {\textbf{Stage 1: AE}\\20 epochs\\Stable data};

% Stage 2
\node[stage, fill=orange!15, right=0.3cm of s1] (s2) {\textbf{Stage 2: NodeFormer}\\60 epochs\\Normal + Event};

% Stage 3
\node[stage, fill=cyan!15, right=0.3cm of s2] (s3) {\textbf{Stage 3: SAC}\\50 epochs\\Learn $\tau$, $\alpha$};

% Stage 4
\node[stage, fill=red!15, right=0.3cm of s3] (s4) {\textbf{Stage 4: Fine-tune}\\20 epochs\\All unfrozen};

% Arrows between stages
\draw[arrow] (s1) -- (s2);
\draw[arrow] (s2) -- (s3);
\draw[arrow] (s3) -- (s4);

% Time arrow below - horizontal line from s1 left edge to s4 right edge
\coordinate (timeline_start) at ([yshift=-0.5cm]s1.south west);
\coordinate (timeline_end) at ([yshift=-0.5cm]s4.south east);
\draw[arrow, gray] (timeline_start) -- (timeline_end) node[midway, below, font=\tiny] {Training Progression};

\end{tikzpicture}%
}
\caption{Multi-stage training pipeline. Stage 1 trains the autoencoder on stable market data. Stage 2 trains both node transformers on their respective data subsets. Stage 3 trains the SAC controller with frozen prediction components. Stage 4 performs end-to-end fine-tuning with all weights unfrozen.}
\label{fig:training_pipeline}
\end{figure}

Figure~\ref{fig:training_pipeline} illustrates the complete multi-stage training pipeline.

The complete training pipeline proceeds in four stages. In Stage 1 (20 epochs), the autoencoder is trained on stable-period data where VIX falls below the 75th percentile of its training-period distribution. Stage 2 (60 epochs) trains both node transformers: the normal pathway on data with low reconstruction error (below the 95th percentile), and the event pathway on high-error data augmented with context features. In Stage 3 (50 epochs), autoencoder and node transformer weights are frozen while the SAC controller learns to optimize the threshold and blending parameters. Finally, Stage 4 (20 epochs) unfreezes all components for end-to-end fine-tuning with reduced learning rates. The architecture is modular by design: Stages 1 and 2 produce a fully functional prediction system in which the routing threshold and blending weight remain at their initialization values ($\tau$ at the 95th percentile of training-set reconstruction errors, $\alpha = 0.5$). Stage 3 adds adaptive control on top of this static configuration, allowing the experimental evaluation to quantify the marginal contribution of the SAC controller by comparing the system with and without it. At inference time, all weights, including the SAC policy network, are frozen. The policy produces state-dependent routing decisions through its fixed learned mapping, with no gradient updates or reward computation during the test period.

\subsection{Loss Functions}

The prediction networks minimize a composite loss:

\begin{equation}
\mathcal{L} = \lambda_1 \mathcal{L}_{\text{MSE}} + \lambda_2 \mathcal{L}_{\text{DIR}} + \lambda_3 \mathcal{L}_{\text{REG}}
\end{equation}

where $\mathcal{L}_{\text{MSE}} = \frac{1}{N} \sum_{i,t,h} (y_{i,t+h} - \hat{y}_{i,t+h})^2$ is the mean squared error between predicted and actual prices. The directional loss $\mathcal{L}_{\text{DIR}}$ is a binary cross-entropy term that explicitly rewards correct prediction of price movement direction, since minimizing magnitude error alone does not guarantee directional accuracy:

\begin{equation}
\mathcal{L}_{\text{DIR}} = -\frac{1}{N} \sum_{i,t,h} \left[ d_{i,t,h} \log p_{i,t,h} + (1 - d_{i,t,h}) \log(1 - p_{i,t,h}) \right]
\end{equation}

where $d_{i,t,h} = \mathbb{I}(y_{i,t+h} > y_{i,t})$ is the true direction indicator and $p_{i,t,h}$ is the predicted probability of a price increase. The regularization term $\mathcal{L}_{\text{REG}} = \|\vect{\theta}\|_2^2$ applies L2 weight decay over all trainable parameters $\vect{\theta}$, penalizing large weight magnitudes to prevent overfitting. Loss weights are $\lambda_1 = 1.0$, $\lambda_2 = 0.5$, $\lambda_3 = 10^{-4}$.

Table~\ref{tab:hyperparameters} summarizes all model hyperparameters across the three components.

\begin{table}[!htbp]
\centering
\caption{Model Hyperparameters}
\label{tab:hyperparameters}
\resizebox{\columnwidth}{!}{%
\begin{tabular}{llc}
\toprule
\textbf{Component} & \textbf{Parameter} & \textbf{Value} \\
\midrule
\multirow{4}{*}{Autoencoder} & Hidden layers & [64, 32] \\
& Latent dimension & 32 \\
& Learning rate & $10^{-3}$ \\
& Training epochs & 20 \\
\midrule
\multirow{7}{*}{Node Transformer} & Layers & 6 \\
& Attention heads & 8 \\
& Model dimension & 512 \\
& FFN dimension & 2048 \\
& Dropout & 0.1 \\
& Learning rate & $10^{-4}$ \\
& Input sequence length & 252 days \\
\midrule
\multirow{5}{*}{SAC Controller} & Hidden layers & [256, 256] \\
& Learning rate & $3 \times 10^{-4}$ \\
& Soft update $\tau$ & 0.005 \\
& Replay buffer & $10^5$ \\
& Training epochs & 50 \\
\midrule
\multirow{3}{*}{Stage 4 Fine-tuning} & AE learning rate & $10^{-4}$ \\
& Node Transformer learning rate & $10^{-5}$ \\
& SAC learning rate & $3 \times 10^{-5}$ \\
\bottomrule
\end{tabular}}
\end{table}

%========================================
% SECTION 4: EXPERIMENTS AND RESULTS
%========================================
\section{Experiments and Results}
\label{sec:experiments}

\subsection{Dataset and Experimental Setup}

The dataset comprises two complementary data streams for 20 S\&P 500 stocks spanning January 1982 to March 2025. The Financial Market Data (FMD) stream consists of daily OHLCV (open, high, low, close, volume) price data sourced from Yahoo Finance, providing adjusted close prices that account for stock splits and dividends. Each trading day produces a five-dimensional price vector per stock alongside the trading volume, from which 11 additional technical indicators are derived (SMA, EMA, RSI, MACD, returns, log returns, and rolling volatility) as described in Section~\ref{sec:methodology}. The sentiment stream draws on two datasets. The first is the Market Sentiment Evaluation (MSE) dataset \cite{cortis2017semeval}, a publicly available corpus of finance-related social media messages annotated by financial experts with sentiment scores in $[-1, +1]$, which serves as ground truth for fine-tuning the BERT sentiment classifier. The second is the Comprehensive Stock Sentiment (CSS) dataset, which was introduced in \cite{alridhawi2025nodeformer} and was constructed using the X (formerly Twitter) API through systematic searches for posts mentioning the 20 stock tickers, yielding approximately 4.2 million posts covering January 2007 to March 2025. The fine-tuned BERT model is applied to the CSS corpus to generate daily sentiment scores for each stock, which are then aggregated and fed into the prediction framework as additional input features.

Table~\ref{tab:stock_universe} lists the complete stock universe. Stocks were selected to span nine distinct sectors, ensuring that the graph structure captures both intra-sector and cross-sector dependencies. The selection also prioritizes variation in market capitalization, trading volume, and volatility characteristics to evaluate robustness across different stock profiles. For companies with IPO dates after 1982 (e.g., Salesforce incorporated 1999, Netflix 2002, Visa 2008), data begins at their first available trading date, and these stocks are included in training only from their listing date onward.

\begin{table}[!htbp]
\centering
\caption{Stock Universe: 20 S\&P 500 Constituents Across 9 Sectors}
\label{tab:stock_universe}
\resizebox{\columnwidth}{!}{%
\begin{tabular}{llc}
\toprule
\textbf{Sector} & \textbf{Stock (Ticker)} & \textbf{Data Start} \\
\midrule
\multirow{3}{*}{Technology} & Apple (AAPL) & 1982 \\
& Microsoft (MSFT) & 1986 \\
& Salesforce (CRM) & 1999 \\
\midrule
\multirow{2}{*}{Financial Services} & JPMorgan Chase (JPM) & 1982 \\
& Visa (V) & 2008 \\
\midrule
\multirow{3}{*}{Healthcare} & Johnson \& Johnson (JNJ) & 1982 \\
& UnitedHealth Group (UNH) & 1984 \\
& Pfizer (PFE) & 1982 \\
\midrule
\multirow{2}{*}{Retail} & Walmart (WMT) & 1982 \\
& Home Depot (HD) & 1982 \\
\midrule
\multirow{2}{*}{Energy} & ExxonMobil (XOM) & 1982 \\
& Chevron (CVX) & 1982 \\
\midrule
\multirow{4}{*}{Consumer Goods} & Procter \& Gamble (PG) & 1982 \\
& Coca-Cola (KO) & 1982 \\
& Nike (NKE) & 1982 \\
& McDonald's (MCD) & 1982 \\
\midrule
Entertainment & Netflix (NFLX) & 2002 \\
\midrule
Telecommunications & Verizon (VZ) & 1982 \\
\midrule
\multirow{2}{*}{Industrials} & Boeing (BA) & 1982 \\
& Caterpillar (CAT) & 1982 \\
\bottomrule
\end{tabular}}
\end{table}

Temporal splits maintain strict chronological separation to prevent any leakage of future information into training. The training set spans January 1982 to December 2010 (approximately 70\% of the temporal range), encompassing multiple market cycles including the 1987 crash, the dot-com bubble and its collapse, and the 2008 financial crisis. The validation set covers January 2011 to December 2016 (approximately 15\%), a period of relatively steady recovery used for hyperparameter tuning and early stopping. The test set spans January 2017 to March 2025 (approximately 15\%), including the 2018 correction, the 2020 COVID crash, and the 2022 market decline, which provide rigorous evaluation under diverse volatility conditions.

Since the X platform (formerly Twitter) was founded in 2006, sentiment data covers January 2007 to March 2025. For the 1982-2006 portion of training, sentiment features are set to zero, meaning the model learns to operate with and without sentiment depending on the data period. During the validation and test periods, full sentiment coverage is available.

\subsection{Evaluation Metrics}

Model performance is assessed using five complementary metrics. The primary metric is Mean Absolute Percentage Error (MAPE), defined as:

\begin{equation}
\text{MAPE} = \frac{100}{N} \sum_{i=1}^{N} \left| \frac{y_i - \hat{y}_i}{y_i} \right|
\end{equation}

MAPE provides intuitive interpretation as percentage deviation from actual prices. As a complementary measure, Root Mean Squared Error (RMSE) penalizes large errors more heavily due to the squaring operation and is computed in normalized price units, where each stock's prices are z-scored individually to enable fair cross-stock aggregation:

\begin{equation}
\text{RMSE} = \sqrt{\frac{1}{N} \sum_{i=1}^{N} (y_i - \hat{y}_i)^2}
\end{equation}

Directional Accuracy (DA) measures the proportion of correctly predicted price movement directions, which is particularly relevant for trading applications where the sign of the predicted move often matters more than its magnitude:

\begin{equation}
\text{DA} = \frac{100}{N} \sum_{i=1}^{N} \mathbb{I}\!\left(\text{sign}(\hat{y}_{i,t+h} - y_{i,t}) = \text{sign}(y_{i,t+h} - y_{i,t})\right)
\end{equation}

Theil's U statistic provides a scale-independent benchmark by comparing forecast error to that of a naive random walk that predicts tomorrow's price as today's price:

\begin{equation}
U = \frac{\sqrt{\sum_{t} (y_{t+1} - \hat{y}_{t+1})^2}}{\sqrt{\sum_{t} (y_{t+1} - y_t)^2}}
\end{equation}

Values of $U < 1$ indicate that the model outperforms the naive baseline, making this metric particularly informative for long time series where absolute price level changes can affect percentage-based measures. Finally, the Confidence Tracking Rate (CTR) captures the proportion of predictions where model confidence (measured as inverse prediction variance across the dual pathway outputs) agrees with actual accuracy:

\begin{equation}
\text{CTR} = \frac{1}{NT} \sum_{i,t} \mathbb{I}\!\left((\text{conf}_{i,t} > \bar{c}) = (|\hat{y}_{i,t+h} - y_{i,t+h}| < \bar{\epsilon})\right)
\end{equation}

where $\text{conf}_{i,t}$ is the inverse prediction variance from the two pathways, $\bar{c}$ is the median confidence, and $\bar{\epsilon}$ is the median absolute error across all predictions. CTR indicates whether the model ``knows when it knows,'' a property valuable for risk-sensitive downstream applications.

\subsection{Baseline Models}

Baselines span statistical methods (ARIMA, VAR, MS-VAR \cite{krolzig1997markov}), classical machine learning (Random Forest, SVR, XGBoost), deep learning (LSTM, Simple Transformer), multimodal and regime-switching approaches (BERT Sentiment + LSTM, HMM-LSTM), recent time-series transformers (TimesNet \cite{wu2023timesnet}, PatchTST \cite{nie2023patchtst}, iTransformer \cite{liu2024itransformer}), and the Integrated NodeFormer-BERT model from prior work \cite{alridhawi2025nodeformer}.

To ensure fair comparison, all baselines share the same experimental conditions wherever the model class permits. Every model uses identical temporal splits (training: 1982--2010, validation: 2011--2016, test: 2017--2025), the same expanding-window z-score normalization described in Section~\ref{sec:methodology}, and the same missing-data imputation strategy. Models capable of multivariate input (Random Forest, SVR, XGBoost, LSTM, Simple Transformer, BERT Sentiment + LSTM, TimesNet, PatchTST, iTransformer, HMM-LSTM, Integrated NodeFormer-BERT) receive the same 17-dimensional feature vector comprising OHLCV data and 11 derived technical indicators. Daily sentiment scores produced by the fine-tuned BERT classifier are appended to the feature set for all multivariate models, so that any advantage from sentiment information is available to baselines as well as to the proposed framework. ARIMA operates on the univariate closing price series for each stock independently, VAR jointly models the closing prices of all 20 stocks, and MS-VAR extends the VAR specification with Markov-switching regime dynamics over the same joint price series, since these statistical methods are not designed to incorporate arbitrary exogenous feature vectors. The target variable for all models is identical: the closing price at horizon $h \in \{1, 5, 20\}$ trading days ahead.

The Simple Transformer baseline uses the same encoder architecture as the proposed node transformer (6 layers, 8 attention heads, 512-dimensional representations) but processes each stock's time series independently without graph structure or inter-stock attention, isolating the contribution of graph-based relational modeling. The BERT Sentiment + LSTM baseline combines the same BERT-derived sentiment scores with a two-layer LSTM through concatenation-based fusion, testing whether the attention-based integration in the proposed architecture provides meaningful improvement over straightforward feature combination. The Integrated NodeFormer-BERT model reproduces our prior work \cite{alridhawi2025nodeformer} with its published hyperparameters, serving as the primary single-pathway baseline against which architectural additions are measured.

PatchTST \cite{nie2023patchtst} segments each stock's multivariate input time series into overlapping patches and applies a transformer encoder with self-attention over the patch sequence, capturing local temporal patterns within patches and long-range dependencies across them. Its channel-independent design processes each feature dimension separately before aggregating predictions, which limits its ability to model cross-feature interactions. iTransformer \cite{liu2024itransformer} inverts the conventional transformer architecture by applying self-attention across the variate (feature) dimension rather than the temporal dimension, enabling it to capture dependencies among price, volume, technical indicators, and sentiment features at each time step. TimesNet \cite{wu2023timesnet} extends temporal modeling by transforming one-dimensional time series into two-dimensional tensors based on learned multi-periodicity structure, applying inception-based convolution blocks to capture both intra-period and inter-period variation. All three recent time-series transformers process each stock's feature set independently without graph structure or cross-stock attention, isolating the contribution of relational modeling in the proposed framework.

The Markov-Switching VAR (MS-VAR) \cite{krolzig1997markov} extends the VAR baseline with $K = 3$ latent regime states governed by a first-order Markov chain, allowing the intercepts and error covariance to vary across regimes while the autoregressive coefficients remain regime-invariant (MSIH specification). Regime transitions are inferred through maximum likelihood estimation of the full joint model, in contrast to our autoencoder-based approach which detects anomalies from reconstruction error without specifying the number or parametric structure of regimes a priori. The HMM-LSTM baseline combines a Hidden Markov Model with $K = 3$ states for regime detection with three regime-specific two-layer LSTMs, each trained on data assigned to its corresponding regime by the Viterbi decoder. At inference, the HMM identifies the most likely current regime and routes the input to the corresponding LSTM, producing a regime-conditional forecast. This architecture provides the most direct comparison to our framework: it replaces the autoencoder with an HMM for regime detection, the node transformers with LSTMs for prediction, and omits adaptive control entirely, using fixed routing with no blending across pathways.

Each baseline underwent hyperparameter tuning via grid search on validation data, with the search ranges and selected values reported in Table~\ref{tab:hp_search} (Appendix).

\subsection{Main Results}

Results are reported for two variants of the proposed framework. The full model (AE-NodeFormer + SAC) includes all components and completes all four training stages. The ablated variant (AE-NodeFormer, no SAC) retains the autoencoder and dual node transformers but removes the reinforcement learning controller entirely, skipping Stage 3 of the training pipeline. In this variant, the routing threshold is fixed at $\tau = e_{95}$, the 95th percentile of training-set reconstruction errors, which is the same initialization used by the full model before SAC adaptation begins. This percentile is a standard choice in anomaly detection, classifying the top 5\% of reconstruction errors as anomalous. The blending weight is held constant at $\alpha = 0.5$, assigning equal contribution to both pathways regardless of market conditions. Comparing the two variants isolates the contribution of adaptive parameter tuning from the architectural benefits of autoencoder routing and dual-pathway specialization.

Table~\ref{tab:results_1day} presents 1-day ahead closing price prediction results across all baselines and proposed variants.

\begin{table}[!htbp]
\centering
\caption{1-Day Ahead Closing Price Prediction Results. Best Results in Bold.}
\label{tab:results_1day}
\resizebox{\columnwidth}{!}{%
\begin{tabular}{lccccc}
\toprule
\textbf{Model} & \textbf{MAPE} & \textbf{RMSE} & \textbf{DA} & \textbf{Theil's U} & \textbf{CTR} \\
\midrule
ARIMA \cite{box1976time} & 1.20\% & 1.35 & 55\% & 0.98 & 51\% \\
VAR \cite{sims1980macroeconomics} & 1.10\% & 1.30 & 56\% & 0.95 & 52\% \\
MS-VAR \cite{krolzig1997markov} & 1.02\% & 1.22 & 57\% & 0.90 & 53\% \\
Random Forest \cite{breiman2001random} & 1.10\% & 1.25 & 57\% & 0.92 & 53\% \\
SVR \cite{vapnik1995nature} & 1.20\% & 1.40 & 54\% & 1.02 & 50\% \\
XGBoost \cite{chen2016xgboost} & 1.00\% & 1.15 & 59\% & 0.85 & 55\% \\
LSTM \cite{hochreiter1997long} & 1.00\% & 1.20 & 58\% & 0.88 & 54\% \\
Simple Transformer \cite{vaswani2017attention} & 0.90\% & 1.10 & 61\% & 0.80 & 57\% \\
BERT Sent. + LSTM \cite{devlin2019bert} & 0.90\% & 1.05 & 62\% & 0.78 & 58\% \\
HMM-LSTM \cite{hamilton1989new} & 0.87\% & 1.02 & 64\% & 0.76 & 60\% \\
TimesNet \cite{wu2023timesnet} & 0.85\% & 1.00 & 63\% & 0.75 & 59\% \\
PatchTST \cite{nie2023patchtst} & 0.83\% & 0.98 & 64\% & 0.74 & 59\% \\
iTransformer \cite{liu2024itransformer} & 0.82\% & 0.97 & 64\% & 0.73 & 61\% \\
Integrated NF-BERT \cite{alridhawi2025nodeformer} & 0.80\% & 0.95 & 65\% & 0.72 & 62\% \\
\midrule
AE-NodeFormer (no SAC) & 0.68\% & 0.88 & 69\% & 0.68 & 64\% \\
\textbf{AE-NodeFormer + SAC} & \textbf{0.59\%} & \textbf{0.82} & \textbf{72\%} & \textbf{0.64} & \textbf{67\%} \\
\bottomrule
\end{tabular}}
\end{table}

The proposed AE-NodeFormer + SAC achieves 0.59\% MAPE, representing a 26\% relative improvement over the Integrated NodeFormer-BERT baseline (0.80\%) and a 28\% improvement over iTransformer (0.82\%), the strongest recent time-series transformer. Directional accuracy reaches 72\%, a 7 percentage point gain over the graph-based baseline. Among regime-switching approaches, HMM-LSTM achieves 0.87\% MAPE, outperforming the basic LSTM (1.00\%) by 13\% through regime-specific specialization, yet still trailing the proposed model by 32\%, indicating that the combination of autoencoder-based anomaly detection, graph-aware dual pathways, and adaptive control provides substantially greater benefit than parametric regime detection with independent LSTMs. The recent time-series transformers (TimesNet 0.85\%, PatchTST 0.83\%, iTransformer 0.82\%) cluster near the Integrated NodeFormer-BERT (0.80\%), confirming that the prior single-pathway architecture was already competitive with current state-of-the-art forecasting models and that the improvements in the present work stem from the regime-aware architectural innovations rather than from a weak baseline. All pairwise improvements of the proposed model over iTransformer, PatchTST, and HMM-LSTM are statistically significant (Diebold-Mariano test \cite{diebold1995comparing}, $p < 0.001$ in each case).

To contextualize the directional accuracy, we computed a naive long-only baseline: predicting ``up'' for every day. Over the 2017-2025 test period, this naive strategy achieves 54\% DA on average across the 20 stocks, reflecting the slight upward drift in equity markets. The 72\% DA of the proposed model thus represents an 18 percentage point improvement over this trivial baseline, confirming that the model captures predictive signal beyond simple market drift.

To assess generalization across forecasting horizons, Table~\ref{tab:results_5day} and Table~\ref{tab:results_20day} present 5-day and 20-day ahead closing price results.

\begin{table}[!htbp]
\centering
\caption{5-Day Ahead Closing Price Prediction Results}
\label{tab:results_5day}
\resizebox{\columnwidth}{!}{%
\begin{tabular}{lccccc}
\toprule
\textbf{Model} & \textbf{MAPE} & \textbf{RMSE} & \textbf{DA} & \textbf{Theil's U} & \textbf{CTR} \\
\midrule
ARIMA \cite{box1976time} & 2.05\% & 2.30 & 51\% & 1.05 & 47\% \\
VAR \cite{sims1980macroeconomics} & 1.88\% & 2.10 & 52\% & 1.00 & 48\% \\
MS-VAR \cite{krolzig1997markov} & 1.70\% & 1.90 & 53\% & 0.95 & 49\% \\
Random Forest \cite{breiman2001random} & 1.92\% & 2.15 & 52\% & 1.02 & 48\% \\
SVR \cite{vapnik1995nature} & 2.10\% & 2.35 & 50\% & 1.08 & 46\% \\
XGBoost \cite{chen2016xgboost} & 1.68\% & 1.88 & 54\% & 0.92 & 50\% \\
LSTM \cite{hochreiter1997long} & 1.65\% & 1.85 & 54\% & 0.93 & 50\% \\
Simple Transformer \cite{vaswani2017attention} & 1.50\% & 1.68 & 56\% & 0.85 & 53\% \\
BERT Sent. + LSTM \cite{devlin2019bert} & 1.48\% & 1.65 & 57\% & 0.83 & 54\% \\
HMM-LSTM \cite{hamilton1989new} & 1.45\% & 1.60 & 59\% & 0.82 & 55\% \\
TimesNet \cite{wu2023timesnet} & 1.40\% & 1.55 & 58\% & 0.81 & 56\% \\
PatchTST \cite{nie2023patchtst} & 1.38\% & 1.52 & 59\% & 0.79 & 55\% \\
iTransformer \cite{liu2024itransformer} & 1.35\% & 1.50 & 59\% & 0.80 & 57\% \\
Integrated NF-BERT \cite{alridhawi2025nodeformer} & 1.30\% & 1.45 & 61\% & 0.78 & 58\% \\
\midrule
AE-NodeFormer (no SAC) & 1.15\% & 1.32 & 64\% & 0.74 & 60\% \\
\textbf{AE-NodeFormer + SAC} & \textbf{1.05\%} & \textbf{1.25} & \textbf{67\%} & \textbf{0.70} & \textbf{63\%} \\
\bottomrule
\end{tabular}}
\end{table}

\begin{table}[!htbp]
\centering
\caption{20-Day Ahead Closing Price Prediction Results}
\label{tab:results_20day}
\resizebox{\columnwidth}{!}{%
\begin{tabular}{lccccc}
\toprule
\textbf{Model} & \textbf{MAPE} & \textbf{RMSE} & \textbf{DA} & \textbf{Theil's U} & \textbf{CTR} \\
\midrule
ARIMA \cite{box1976time} & 3.10\% & 3.45 & 48\% & 1.12 & 44\% \\
VAR \cite{sims1980macroeconomics} & 2.85\% & 3.20 & 49\% & 1.05 & 45\% \\
MS-VAR \cite{krolzig1997markov} & 2.55\% & 2.85 & 50\% & 0.98 & 46\% \\
Random Forest \cite{breiman2001random} & 2.90\% & 3.25 & 49\% & 1.08 & 45\% \\
SVR \cite{vapnik1995nature} & 3.20\% & 3.55 & 47\% & 1.15 & 43\% \\
XGBoost \cite{chen2016xgboost} & 2.60\% & 2.90 & 51\% & 0.98 & 47\% \\
LSTM \cite{hochreiter1997long} & 2.50\% & 2.80 & 51\% & 0.96 & 47\% \\
Simple Transformer \cite{vaswani2017attention} & 2.25\% & 2.52 & 53\% & 0.90 & 50\% \\
BERT Sent. + LSTM \cite{devlin2019bert} & 2.20\% & 2.45 & 54\% & 0.88 & 51\% \\
HMM-LSTM \cite{hamilton1989new} & 2.12\% & 2.35 & 55\% & 0.87 & 52\% \\
TimesNet \cite{wu2023timesnet} & 2.08\% & 2.30 & 56\% & 0.85 & 52\% \\
PatchTST \cite{nie2023patchtst} & 2.05\% & 2.28 & 55\% & 0.84 & 53\% \\
iTransformer \cite{liu2024itransformer} & 2.00\% & 2.22 & 56\% & 0.86 & 53\% \\
Integrated NF-BERT \cite{alridhawi2025nodeformer} & 1.90\% & 2.15 & 57\% & 0.85 & 54\% \\
\midrule
AE-NodeFormer (no SAC) & 1.70\% & 2.00 & 60\% & 0.82 & 56\% \\
\textbf{AE-NodeFormer + SAC} & \textbf{1.55\%} & \textbf{1.85} & \textbf{63\%} & \textbf{0.78} & \textbf{59\%} \\
\bottomrule
\end{tabular}}
\end{table}

Performance improvements persist across all prediction horizons. At the 5-day horizon, the proposed model achieves 1.05\% MAPE compared to 1.30\% for the Integrated NodeFormer-BERT and 1.35\% for iTransformer, maintaining a 19\% and 22\% relative advantage respectively. At 20 days, these gaps widen further: the proposed model reaches 1.55\% MAPE versus 1.90\% for the graph-based baseline and 2.00\% for iTransformer, reflecting the increasing value of regime-aware routing as the prediction horizon extends and structural regime shifts become more consequential. Several statistical and classical ML baselines (ARIMA, VAR, Random Forest, SVR) produce Theil's U values exceeding 1.0 at the 20-day horizon, indicating that they underperform the naive random walk at longer horizons, a well-known limitation of models without explicit temporal or regime-adaptive structure. In contrast, all transformer-based and regime-switching models maintain Theil's U below 1.0 across all horizons. Directional accuracy for the proposed model declines from 72\% at 1-day to 63\% at 20-day, a more gradual degradation than iTransformer (64\% to 56\%) or HMM-LSTM (64\% to 55\%), suggesting that the combination of autoencoder routing and SAC adaptation captures structural signals that remain informative beyond short-term momentum.

\subsection{Per-Stock Results}

To examine cross-sectional variation, Table~\ref{tab:per_stock} presents 1-day ahead closing price results for all 20 stocks in the universe, grouped by sector.

\begin{table}[!htbp]
\centering
\caption{Per-Stock 1-Day Ahead Closing Price Results (AE-NodeFormer + SAC)}
\label{tab:per_stock}
\resizebox{\columnwidth}{!}{%
\begin{tabular}{llcccc}
\toprule
\textbf{Sector} & \textbf{Stock} & \textbf{MAPE} & \textbf{RMSE} & \textbf{DA} & \textbf{Theil's U} \\
\midrule
\multirow{3}{*}{Technology} & AAPL & 0.62\% & 0.88 & 69\% & 0.68 \\
& MSFT & 0.50\% & 0.72 & 74\% & 0.60 \\
& CRM & 0.63\% & 0.89 & 70\% & 0.67 \\
\midrule
\multirow{2}{*}{Financial Services} & JPM & 0.70\% & 1.02 & 67\% & 0.72 \\
& V & 0.48\% & 0.69 & 76\% & 0.58 \\
\midrule
\multirow{3}{*}{Healthcare} & JNJ & 0.44\% & 0.64 & 77\% & 0.55 \\
& UNH & 0.52\% & 0.74 & 73\% & 0.61 \\
& PFE & 0.58\% & 0.80 & 72\% & 0.64 \\
\midrule
\multirow{2}{*}{Retail} & WMT & 0.42\% & 0.62 & 78\% & 0.54 \\
& HD & 0.48\% & 0.68 & 75\% & 0.59 \\
\midrule
\multirow{2}{*}{Energy} & XOM & 1.10\% & 1.38 & 63\% & 0.82 \\
& CVX & 0.95\% & 1.22 & 64\% & 0.79 \\
\midrule
\multirow{4}{*}{Consumer Goods} & PG & 0.43\% & 0.62 & 77\% & 0.55 \\
& KO & 0.44\% & 0.65 & 76\% & 0.56 \\
& NKE & 0.56\% & 0.73 & 72\% & 0.63 \\
& MCD & 0.45\% & 0.65 & 77\% & 0.56 \\
\midrule
Entertainment & NFLX & 0.75\% & 1.05 & 66\% & 0.75 \\
\midrule
Telecommunications & VZ & 0.47\% & 0.67 & 75\% & 0.58 \\
\midrule
\multirow{2}{*}{Industrials} & BA & 0.68\% & 0.94 & 68\% & 0.71 \\
& CAT & 0.60\% & 0.81 & 71\% & 0.66 \\
\midrule
\multicolumn{2}{l}{\textbf{Mean}} & \textbf{0.59\%} & \textbf{0.82} & \textbf{72\%} & \textbf{0.64} \\
\bottomrule
\end{tabular}}
\end{table}

Individual stock performance spans from 0.42\% MAPE (WMT) to 1.10\% (XOM), with 16 of 20 stocks falling below 0.70\%. The error distribution aligns with established differences in equity predictability. Defensive stocks with stable revenue profiles, including WMT (0.42\%), PG (0.43\%), JNJ (0.44\%), KO (0.44\%), and MCD (0.45\%), cluster at the low end regardless of sector classification, suggesting that the predictability advantage stems from fundamental business stability rather than broad sectoral factors. Energy stocks occupy the highest error positions, with both XOM (1.10\%) and CVX (0.95\%) exhibiting MAPE values roughly double the universe mean, consistent with the dominant influence of exogenous commodity price movements that the autoencoder's feature-based reconstruction cannot fully anticipate. Within sectors, meaningful variation persists: in healthcare, JNJ (0.44\%) substantially outperforms PFE (0.58\%), plausibly reflecting Pfizer's heightened pipeline-driven volatility during the test period; in technology, MSFT (0.50\%) outperforms both AAPL (0.62\%) and CRM (0.63\%), consistent with differences in revenue diversification and product-cycle exposure. Financial services exhibit a similar spread, where Visa's stable payment-processing model yields considerably lower error (0.48\%) than JPMorgan's sensitivity to interest rate and credit dynamics (0.70\%). Although the sample of two to four stocks per sector does not support formal statistical claims about sectoral predictability, the consistency of observed patterns, with both energy stocks at the top of the error distribution and five defensive consumer and healthcare names clustered near the bottom, suggests that stock-level characteristics such as earnings stability, commodity exposure, and idiosyncratic volatility interact with the regime detection mechanism in interpretable ways. Theil's U remains below 1.0 for all 20 stocks without exception, confirming that the model outperforms the naive random-walk baseline across the full predictability spectrum. Directional accuracy ranges from 63\% (XOM) to 78\% (WMT), with every stock exceeding the 54\% naive long-only baseline reported in the main results, indicating that the regime-aware routing mechanism provides meaningful predictive signal even for the most volatile equities in the universe.

\subsection{Ablation Study}

To quantify the contribution of each architectural component, Table~\ref{tab:ablation} reports results from systematically removing one component at a time, with all other components held constant or adapted to the reduced architecture as described below.

The \textit{No SAC} configuration removes the reinforcement learning controller entirely, skipping Stage 3 of the training pipeline. The autoencoder and dual node transformers retain their Stage 1 and Stage 2 trained weights. The routing threshold is fixed at $\tau = e_{95}$, the 95th percentile of training-set reconstruction errors, and the blending weight is held at $\alpha = 0.5$, assigning equal contribution to both pathways regardless of market conditions. This variant isolates the benefit of adaptive parameter tuning from the architectural contributions of regime-aware routing and pathway specialization.

The \textit{No Dual Paths} configuration replaces the two specialized node transformers with a single node transformer that processes all data regardless of regime classification. The single pathway retains the same architectural hyperparameters as each individual pathway in the full model (6 layers, 8 attention heads, 512 model dimension), so that any performance difference reflects the architectural choice of pathway specialization rather than a difference in model capacity. The autoencoder is retained, and its reconstruction error $e_t$ together with a binary regime indicator (determined by threshold $\tau$) are concatenated to the single pathway's input feature vector, providing regime context without architectural separation. The SAC controller remains active but operates over a reduced action space: it adjusts only $\tau$ to optimize the anomaly detection threshold, since the blending weight $\alpha$ is undefined when a single pathway produces the output. This variant quantifies the value of allocating independent representational capacity to normal and anomalous conditions, as opposed to conditioning a shared pathway on a regime signal.

The \textit{No AE} configuration removes the autoencoder, which cascades into removing dual-pathway routing and the SAC controller, since both depend on the reconstruction error that the autoencoder produces. The resulting system is a single node transformer processing the standard feature set augmented with BERT sentiment scores, architecturally equivalent to the Integrated NodeFormer-BERT baseline from prior work \cite{alridhawi2025nodeformer}. This configuration serves as the reference point from which the incremental contributions of regime-aware routing, pathway specialization, and adaptive control are jointly measured.

\begin{table}[!htbp]
\centering
\caption{Ablation Study: 1-Day MAPE}
\label{tab:ablation}
\resizebox{\columnwidth}{!}{%
\begin{tabular}{lcc}
\toprule
\textbf{Configuration} & \textbf{MAPE} & \textbf{$\Delta$ vs Full} \\
\midrule
Full Model (AE + Dual NF + SAC) & 0.59\% & -- \\
No SAC (AE + Dual NF) & 0.68\% & +15.3\% \\
No Dual Paths (AE + Single NF + SAC) & 0.63\% & +6.8\% \\
No AE (Single NF + BERT, baseline) & 0.80\% & +35.6\% \\
\bottomrule
\end{tabular}}
\end{table}

Autoencoder routing contributes most substantially, with its removal increasing MAPE by 35.6\% relative. This large degradation reflects the fact that the autoencoder provides the foundational regime signal upon which both pathway routing and adaptive control depend; its removal eliminates the entire regime-aware processing chain. The SAC controller contributes the next largest improvement at 15.3\%, confirming that adaptive tuning of $\tau$ and $\alpha$ based on prediction feedback materially outperforms static initialization, particularly during regime transitions where the optimal threshold and blending weight shift over time. The dual-pathway architecture contributes a smaller but meaningful 6.8\% improvement, indicating that allocating independent weights to normal and anomalous conditions yields better representations than conditioning a single pathway on a binary regime indicator, even when both configurations receive the same regime information from the autoencoder. All components provide statistically significant gains ($p < 0.01$ via paired t-tests across stock-day predictions).

\subsection{Volatility Regime Analysis}

To evaluate regime-specific performance, Table~\ref{tab:volatility} disaggregates MAPE by VIX regime.

\begin{table}[!htbp]
\centering
\caption{1-Day MAPE by Volatility Regime}
\label{tab:volatility}
\resizebox{\columnwidth}{!}{%
\begin{tabular}{lccc}
\toprule
\textbf{Model} & \textbf{Low VIX} & \textbf{Medium VIX} & \textbf{High VIX} \\
\midrule
iTransformer & 0.72\% & 0.92\% & 1.42\% \\
HMM-LSTM & 0.78\% & 0.95\% & 1.35\% \\
Integrated NodeFormer-BERT & 0.70\% & 0.90\% & 1.50\% \\
AE-NodeFormer (no SAC) & 0.60\% & 0.75\% & 1.10\% \\
AE-NodeFormer + SAC & \textbf{0.52\%} & \textbf{0.65\%} & \textbf{0.85\%} \\
\bottomrule
\end{tabular}}
\end{table}

The regime-specific results reveal an instructive pattern. During low-VIX periods, iTransformer (0.72\%) slightly outperforms HMM-LSTM (0.78\%), as the superior representational capacity of the transformer architecture dominates when market dynamics are stable and regime detection adds limited value. During high-VIX periods, this relationship reverses: HMM-LSTM (1.35\%) outperforms iTransformer (1.42\%) because its regime-specific LSTMs adapt to volatile conditions even though its overall architecture is less expressive. The proposed model outperforms both across all VIX levels, maintaining MAPE at 0.85\% during high-volatility periods where iTransformer reaches 1.42\% and the Integrated NodeFormer-BERT baseline reaches 1.50\%. The 40\% relative improvement over iTransformer in high-VIX conditions, compared to 28\% overall, confirms that regime-aware processing is most valuable in precisely the conditions where accurate forecasts matter most for risk management.

\subsection{SAC Controller Behavior}

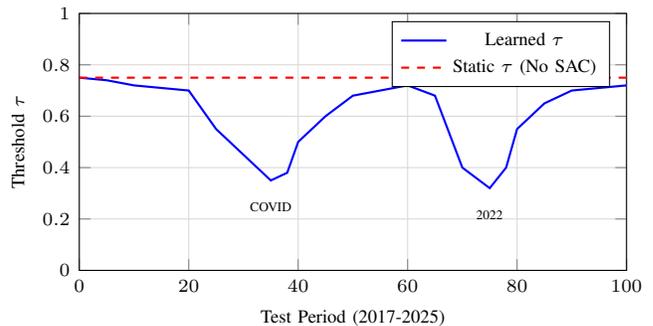
\begin{figure}[!htbp]
\centering
\begin{tikzpicture}
\begin{axis}[
    width=\columnwidth,
    height=5cm,
    xlabel={Test Period (2017-2025)},
    ylabel={Threshold $\tau$},
    xmin=0, xmax=100,
    ymin=0, ymax=1,
    legend pos=north east,
    font=\scriptsize,
    grid=major,
    grid style={gray!30}
]
\addplot[blue, thick, mark=none] coordinates {
    (0, 0.75) (5, 0.74) (10, 0.72) (15, 0.71) (20, 0.70) (25, 0.55) (30, 0.45) (35, 0.35) (38, 0.38) (40, 0.50)
    (45, 0.60) (50, 0.68) (55, 0.70) (60, 0.72) (65, 0.68) (70, 0.40) (75, 0.32) (78, 0.40) (80, 0.55) (85, 0.65) (90, 0.70) (95, 0.71) (100, 0.72)
};
\addlegendentry{Learned $\tau$}
\addplot[red, dashed, thick, mark=none] coordinates {
    (0, 0.75) (100, 0.75)
};
\addlegendentry{Static $\tau$ (No SAC)}

% Annotations for volatile periods - positioned below the line
\node[font=\tiny, anchor=north] at (axis cs:35,0.30) {COVID};
\node[font=\tiny, anchor=north] at (axis cs:75,0.27) {2022};

\end{axis}
\end{tikzpicture}
\caption{Threshold $\tau$ produced by the frozen SAC policy over the test period. Dips correspond to volatile periods (COVID crash around index 35, 2022 market decline around index 75) where elevated reconstruction errors cause the fixed policy to output lower threshold values, routing more data through the event pathway. The dashed red line shows the static threshold ($\tau = e_{95}$) used in the No SAC ablation variant. All policy weights are frozen after training; the trajectory reflects state-dependent outputs from a fixed function, not online learning.}
\label{fig:sac_threshold}
\end{figure}

Because the threshold $\tau$ and blending weight $\alpha$ vary across the test period (Figure~\ref{fig:sac_threshold}), it is important to specify precisely what occurs at inference time and to address the resulting implications for baseline comparability. After Stage 4 training completes, all model weights are frozen, including the SAC actor and critic networks. During testing, the actor network operates as a fixed deterministic function: given the current state $s_t$, it outputs $[\Delta\tau, \Delta\alpha]$ through a single forward pass with no gradient computation, no reward evaluation, and no parameter update. The threshold and blending weight evolve across the test period because the \textit{inputs} to this fixed function change (reconstruction errors rise during volatile markets, recent prediction metrics shift), not because the policy itself is modified. In this respect, the SAC policy at inference time is functionally equivalent to any other feedforward neural network applied to streaming data: its parameters are static, but its outputs depend on input features that vary over time.

This state-dependent inference behavior is not unique to the proposed framework. The HMM-LSTM baseline performs analogous adaptive routing at test time: at each step, the HMM's forward algorithm computes regime posterior probabilities using fixed transition and emission parameters learned during training, and these probabilities determine which regime-specific LSTM produces the forecast. The MS-VAR baseline similarly infers time-varying regime probabilities through its fixed Markov-switching parameters, adjusting intercepts and error covariance accordingly. In all three cases (the proposed SAC policy, the HMM, and the Markov-switching model), a frozen statistical or neural model produces time-varying routing decisions from fixed parameters applied to changing inputs. The proposed framework thus does not enjoy an online learning advantage relative to the regime-switching baselines; rather, the three approaches represent alternative designs for the same underlying capability of state-dependent inference-time adaptation.

A separate question is whether state-dependent routing confers an advantage over the purely static baselines (ARIMA, Random Forest, LSTM, Simple Transformer, and others) that apply fixed parameters uniformly across all market conditions. It does, and this advantage is by design: the central thesis of this work is that regime-aware processing improves prediction quality. Crucially, the ablation study (Table~\ref{tab:ablation}) demonstrates that this advantage does not depend on the SAC controller. The AE-NodeFormer variant without SAC uses entirely static routing parameters ($\tau = e_{95}$, $\alpha = 0.5$) and achieves 0.68\% MAPE, which already outperforms every baseline including iTransformer (0.82\%) and the Integrated NodeFormer-BERT (0.80\%). The architectural contributions of autoencoder-based regime detection and dual-pathway specialization account for the majority of the improvement, with the SAC controller providing an additional 15.3\% relative gain through its state-dependent refinement of routing decisions. The comparison between the static No SAC variant and the baselines is therefore on equal footing with respect to inference-time adaptation.

Regarding the state features themselves, the previous-day prediction metrics ($\text{RMSE}_{t-1}$, $\text{DA}_{t-1}$) included in the SAC state vector are computed by comparing the model's day-$(t-1)$ forecast with the realized closing price, which is publicly available at market open on day $t$. This introduces no information leakage: any practitioner would know whether yesterday's prediction was accurate before making today's forecast. The remaining state features (current reconstruction error $e_t$, recent error history $\bar{e}_{t-k:t}$, and market volatility $\sigma_t$) are computed entirely from the model's own outputs and observable market data, with no access to future prices.

Figure~\ref{fig:sac_threshold} illustrates the threshold trajectory produced by the frozen policy. During stable periods, the policy maps low reconstruction errors to higher thresholds, routing most data through the normal pathway. When volatility increases and reconstruction errors rise, the same fixed policy maps these elevated states to lower thresholds, directing more data to the event pathway. The blending weight $\alpha$ follows a complementary pattern, reducing normal pathway contribution during detected anomalies.

The threshold trajectory reveals interpretable behavior. During the pre-COVID period (2017-2019), the frozen policy maps the prevailing low reconstruction errors to a relatively high threshold ($\tau \approx 0.70$-$0.75$), routing the majority of data through the normal pathway. As the COVID crash unfolds in early 2020, the spike in reconstruction errors causes the policy to output large negative $\Delta\tau$ adjustments, dropping the threshold sharply to approximately 0.35 and activating the event pathway for most stocks. Recovery is gradual: as reconstruction errors slowly normalize, the policy produces small positive adjustments that return the threshold to pre-crisis levels over several months rather than snapping back, reflecting the state-to-action mapping learned during training between persistent elevated errors and cautious threshold recovery. A similar but less severe pattern occurs during the 2022 market decline. Importantly, the policy's mapping was learned entirely from training-period data; no crisis labels, test-period supervision, or weight updates inform the threshold trajectory shown in the figure. The frozen policy generalizes its learned associations between reconstruction error patterns and routing decisions to market events it has never encountered.

\subsection{Statistical Significance}

To confirm that the observed improvements are not attributable to chance, Table~\ref{tab:significance} presents paired t-test results comparing daily squared errors.

\begin{table}[!htbp]
\centering
\caption{Statistical Significance ($n = 1{,}580$ Test Days)}
\label{tab:significance}
\resizebox{\columnwidth}{!}{%
\begin{tabular}{lccc}
\toprule
\textbf{Comparison} & \textbf{$t$-statistic} & \textbf{$p$-value} & \textbf{Cohen's $d$} \\
\midrule
AE-NF+SAC vs Integrated NF-BERT & $-5.82$ & $< 0.0001$ & 0.46 \\
AE-NF+SAC vs AE-NF (no SAC) & $-3.45$ & $0.0006$ & 0.27 \\
AE-NF+SAC vs LSTM & $-7.21$ & $< 0.0001$ & 0.57 \\
\bottomrule
\end{tabular}}
\end{table}

All comparisons achieve $p < 0.001$, with effect sizes (Cohen's $d$) ranging from 0.27 to 0.57 and indicating medium practical significance. The largest effect size (0.57) appears in the comparison against LSTM, which lacks both graph structure and regime awareness. The smallest effect size (0.27) is between the full model and the non-adaptive AE-NodeFormer variant, consistent with the SAC controller's contribution being a refinement over an already strong base architecture rather than a wholesale improvement.

%========================================
% SECTION 5: DISCUSSION
%========================================
\section{Discussion}
\label{sec:discussion}

\subsection{Interpretation of Results}

Regime-aware prediction with adaptive control outperforms homogeneous approaches across all metrics. The 26\% MAPE improvement over the baseline integrated model (0.59\% vs 0.80\%) reflects gains from three sources: regime detection (autoencoder), specialized processing (dual node transformers), and adaptive parameter tuning (SAC controller).

Without requiring explicit per-sample anomaly labels, the autoencoder identifies market states that deviate from normal patterns. High reconstruction errors coincide with periods of elevated volatility, earnings announcements, and macroeconomic shocks, allowing the system to detect anomalies as deviations from its learned representation of typical behavior.

The SAC controller learns to adjust the detection threshold based on prediction outcomes, maintaining a higher threshold during stable periods to prevent unnecessary routing to the event pathway and lowering it during genuine regime shifts to engage event-aware processing. This adaptive behavior emerges from the reward signal rather than hand-crafted rules.

The dual-path architecture enables specialization: the normal pathway develops representations optimized for stable conditions where fundamental factors dominate, while the event pathway incorporates additional context (sentiment spikes, volatility regimes, event characterization) that proves informative during turbulent periods but might add noise during normal conditions.

\subsection{Economic Interpretation}

While the performance improvements carry practical implications, they should be interpreted cautiously. A 7 percentage point improvement in directional accuracy (72\% vs 65\%) carries practical value for trading decisions, and the 43\% relative improvement during high-volatility periods is particularly valuable since these are precisely the conditions where accurate forecasts matter most for risk management.

Transaction costs, market impact, and execution constraints would reduce realized gains from any trading strategy based on these predictions. The economic significance analysis in prior work \cite{alridhawi2025nodeformer} demonstrated that even the baseline model's predictions, when combined with simple trading rules, generate returns exceeding buy-and-hold benchmarks before costs. The improvements documented here should amplify these returns proportionally, though the gap would narrow after incorporating realistic trading frictions.

\subsection{Limitations}

The most consequential limitation concerns data selection. The 20 stocks used in this study are drawn from the current S\&P 500 universe, which introduces survivorship bias: companies that failed, were acquired, or delisted between 1982 and 2025 are absent from the dataset. Because the selected stocks are disproportionately successful over the evaluation period, predictability estimates may be inflated relative to what a real-time investor would experience when choosing from the full market. Performance on a point-in-time universe constructed from historical index constituents could differ, and the reported results should therefore be interpreted as evidence of architectural capability rather than guaranteed trading performance.

Several data-related concerns compound this issue. Edge weights in the graph are initialized from correlations computed over the training period (1982-2010), but market correlations are non-stationary, and relationships that held during this window may weaken or reverse by the test period (2017-2025). The learnable edge refinement mechanism partially compensates by adapting weights during training, though fundamental correlation regime shifts could still affect generalization. Sentiment data from X (formerly Twitter) is available only from 2007 onward; for the 1982-2006 portion of training, sentiment features are set to zero. This means the sentiment modality effectively ``turns on'' partway through training, potentially limiting the model's ability to learn robust sentiment-price relationships from the earlier decades.

On the modeling side, the reinforcement learning controller requires careful tuning of reward weights, temperature, and network architecture, and suboptimal configurations could lead to unstable threshold behavior or poor convergence. The system also learns regime boundaries from its own prediction performance, which risks feedback loops where poor initial predictions lead to suboptimal threshold learning; the staged training protocol mitigates this by pre-training each component before SAC optimization, but the circularity is not fully eliminated. The three-component architecture increases computational requirements, with training time approximately 33\% longer than the baseline model, which may constrain real-time deployment. Although the SAC policy weights are frozen at inference time, the state-dependent routing decisions produced by the fixed policy constitute a form of adaptive processing that purely static baselines (ARIMA, Random Forest, LSTM) do not possess. The regime-switching baselines (HMM-LSTM, MS-VAR) share this property, and the ablation study confirms that the static No SAC variant already outperforms all baselines, but the additional gain from state-dependent threshold tuning should be interpreted as an architectural advantage of the framework rather than an improvement attributable solely to prediction accuracy.

From an economic standpoint, the analysis excludes transaction costs, market impact, and execution constraints. For strategies involving frequent rebalancing, these frictions would reduce realized returns. Backtesting is performed on the same 20-stock universe used for model development, leaving external validity on unseen stocks untested. Finally, sentiment data was collected via the X (formerly Twitter) API, which has undergone significant access policy changes, making exact replication of the sentiment component challenging with current API limitations.

\subsection{Future Directions}

Several extensions could address the limitations identified above. Expanding the stock universe to include historical index constituents would mitigate survivorship bias. Automated SAC hyperparameter tuning via meta-learning could reduce configuration sensitivity. More efficient architectures would enable real-time deployment, and strictly online regime detection without training-period VIX percentiles would eliminate any residual look-ahead.

Beyond addressing limitations, incorporating multiple autoencoders for different anomaly types (e.g., separating liquidity crises from earnings shocks) could refine regime classification. Extending the framework to portfolio optimization, where regime-aware allocation could improve risk-adjusted returns, represents a natural application. Transfer learning to adapt the system to new markets or asset classes without full retraining would broaden practical applicability.

%========================================
% SECTION 6: CONCLUSION
%========================================
\section{Conclusion}
\label{sec:conclusion}

This paper introduced an adaptive framework for stock price prediction that automatically detects market regimes and adjusts processing accordingly. The architecture combines an autoencoder for regime detection, dual node transformer networks specialized for stable and volatile conditions, and a Soft Actor-Critic reinforcement learning controller that learns adaptive regime thresholds from prediction performance.

Experiments on 20 S\&P 500 stocks spanning 1982-2025 demonstrate substantial improvements over prior approaches: the complete system achieves 0.59\% MAPE for one-day predictions compared to 0.80\% for the baseline integrated node transformer, while directional accuracy reaches 72\%, a 7 percentage point improvement. These gains persist across prediction horizons and are most pronounced during volatile periods where the baseline struggles.

The key conceptual contribution is the adaptive learning of regime boundaries. Rather than relying solely on fixed hand-crafted anomaly definitions, the system discovers useful boundaries by optimizing downstream prediction accuracy. This approach avoids the staleness problem of hand-crafted regime rules and the labeling burden of supervised regime classification.

Future work should validate the framework on broader universes, develop more efficient implementations for real-time deployment, and explore extensions to portfolio optimization and risk management applications.

%========================================
% APPENDIX
%========================================
\appendix
\section{Hyperparameter Search Ranges}
\label{sec:appendix}

Table~\ref{tab:hp_search} reports the hyperparameter search space explored for each baseline model. All searches were conducted via grid search on the validation set (2011--2016), with the configuration yielding the lowest validation MAPE selected for test evaluation.

\begin{table}[!htbp]
\centering
\caption{Hyperparameter Search Ranges and Selected Values for Baseline Models}
\label{tab:hp_search}
\resizebox{\columnwidth}{!}{%
\renewcommand{\arraystretch}{0.85}
\begin{tabular}{llll}
\toprule
\textbf{Model} & \textbf{Hyperparameter} & \textbf{Search Range} & \textbf{Selected} \\
\midrule
\multirow{4}{*}{ARIMA}
& AR order ($p$) & $\{0, 1, 2, 3, 4, 5\}$ & Per stock (AIC) \\
& Differencing ($d$) & $\{0, 1, 2\}$ & Per stock (AIC) \\
& MA order ($q$) & $\{0, 1, 2, 3, 4, 5\}$ & Per stock (AIC) \\
& Selection criterion & AIC, BIC & AIC \\
\midrule
\multirow{3}{*}{VAR}
& Lag order & $\{1, 2, 3, \ldots, 10\}$ & 3 \\
& Trend & $\{\text{none, constant, both}\}$ & constant \\
& Selection criterion & AIC, BIC & BIC \\
\midrule
\multirow{5}{*}{Random Forest}
& Number of estimators & $\{100, 200, 500\}$ & 200 \\
& Maximum depth & $\{5, 10, 15, 20, \text{None}\}$ & 15 \\
& Min samples split & $\{2, 5, 10\}$ & 5 \\
& Min samples leaf & $\{1, 2, 4\}$ & 2 \\
& Max features & $\{\sqrt{d},\; \log_2(d),\; 0.5\}$ & $\sqrt{d}$ \\
\midrule
\multirow{4}{*}{SVR}
& Kernel & RBF (fixed) & RBF \\
& Regularization ($C$) & $\{0.1, 1, 10, 100\}$ & 10 \\
& Kernel width ($\gamma$) & $\{\text{scale}, 0.01, 0.1\}$ & scale \\
& Epsilon ($\epsilon$) & $\{0.01, 0.05, 0.1\}$ & 0.05 \\
\midrule
\multirow{8}{*}{XGBoost}
& Number of estimators & $\{100, 500, 1000\}$ & 500 \\
& Maximum depth & $\{3, 5, 7, 10\}$ & 7 \\
& Learning rate & $\{0.01, 0.05, 0.1\}$ & 0.05 \\
& Subsample ratio & $\{0.7, 0.8, 0.9, 1.0\}$ & 0.8 \\
& Column sample ratio & $\{0.7, 0.8, 0.9, 1.0\}$ & 0.8 \\
& Min child weight & $\{1, 3, 5\}$ & 3 \\
& L1 regularization ($\alpha$) & $\{0, 0.01, 0.1\}$ & 0.01 \\
& L2 regularization ($\lambda$) & $\{1, 1.5, 2\}$ & 1.5 \\
\midrule
\multirow{6}{*}{LSTM}
& Hidden dimension & $\{64, 128, 256, 512\}$ & 256 \\
& Number of layers & $\{1, 2, 3\}$ & 2 \\
& Dropout & $\{0.1, 0.2, 0.3\}$ & 0.2 \\
& Learning rate & $\{10^{-4}, 5 \times 10^{-4}, 10^{-3}\}$ & $5 \times 10^{-4}$ \\
& Batch size & $\{32, 64\}$ & 64 \\
& Sequence length & 252 (fixed) & 252 \\
\midrule
\multirow{6}{*}{\shortstack[l]{Simple\\Transformer}}
& Layers & $\{4, 6, 8\}$ & 6 \\
& Attention heads & $\{4, 8\}$ & 8 \\
& Model dimension & $\{256, 512\}$ & 512 \\
& FFN dimension & $\{1024, 2048\}$ & 2048 \\
& Dropout & $\{0.1, 0.2\}$ & 0.1 \\
& Learning rate & $\{10^{-4}, 5 \times 10^{-4}, 10^{-3}\}$ & $10^{-4}$ \\
\midrule
\multirow{5}{*}{\shortstack[l]{BERT Sentiment\\+ LSTM}}
& LSTM hidden dimension & $\{128, 256, 512\}$ & 256 \\
& LSTM layers & $\{1, 2\}$ & 2 \\
& Sentiment fusion & $\{\text{concat., gated}\}$ & concatenation \\
& Dropout & $\{0.1, 0.2, 0.3\}$ & 0.2 \\
& Learning rate & $\{10^{-4}, 5 \times 10^{-4}\}$ & $10^{-4}$ \\
\midrule
\multirow{4}{*}{MS-VAR}
& Number of regimes ($K$) & $\{2, 3, 4\}$ & 3 \\
& Lag order & $\{1, 2, 3, \ldots, 10\}$ & 3 \\
& Switching specification & $\{\text{MSI, MSM, MSIH}\}$ & MSIH \\
& EM convergence tolerance & $\{10^{-6}, 10^{-8}\}$ & $10^{-8}$ \\
\midrule
\multirow{7}{*}{HMM-LSTM}
& HMM states ($K$) & $\{2, 3, 4\}$ & 3 \\
& HMM covariance & $\{\text{diag., full}\}$ & diagonal \\
& LSTM hidden dimension & $\{128, 256, 512\}$ & 256 \\
& LSTM layers & $\{1, 2, 3\}$ & 2 \\
& Dropout & $\{0.1, 0.2, 0.3\}$ & 0.2 \\
& Learning rate & $\{10^{-4}, 5 \times 10^{-4}, 10^{-3}\}$ & $5 \times 10^{-4}$ \\
& Sequence length & 252 (fixed) & 252 \\
\midrule
\multirow{7}{*}{TimesNet}
& Layers & $\{2, 3, 4\}$ & 3 \\
& Model dimension & $\{32, 64, 128\}$ & 64 \\
& Top-$k$ periods & $\{3, 5, 7\}$ & 5 \\
& FFN dimension & $\{64, 128, 256\}$ & 128 \\
& Dropout & $\{0.1, 0.2, 0.3\}$ & 0.2 \\
& Learning rate & $\{10^{-4}, 5 \times 10^{-4}, 10^{-3}\}$ & $10^{-4}$ \\
& Sequence length & 252 (fixed) & 252 \\
\midrule
\multirow{7}{*}{PatchTST}
& Patch length & $\{12, 16, 24\}$ & 16 \\
& Stride & $\{8, 12, 16\}$ & 8 \\
& Layers & $\{3, 4, 6\}$ & 4 \\
& Attention heads & $\{4, 8\}$ & 8 \\
& Model dimension & $\{128, 256, 512\}$ & 256 \\
& Dropout & $\{0.1, 0.2, 0.3\}$ & 0.2 \\
& Learning rate & $\{10^{-4}, 5 \times 10^{-4}, 10^{-3}\}$ & $10^{-4}$ \\
\midrule
\multirow{6}{*}{iTransformer}
& Layers & $\{3, 4, 6\}$ & 4 \\
& Attention heads & $\{4, 8\}$ & 8 \\
& Model dimension & $\{128, 256, 512\}$ & 256 \\
& FFN dimension & $\{256, 512, 1024\}$ & 512 \\
& Dropout & $\{0.1, 0.2\}$ & 0.1 \\
& Learning rate & $\{10^{-4}, 5 \times 10^{-4}, 10^{-3}\}$ & $10^{-4}$ \\
\midrule
Integrated NF-BERT
& \multicolumn{3}{l}{Architecture and hyperparameters fixed per \cite{alridhawi2025nodeformer}.} \\
\bottomrule
\end{tabular}}
\end{table}

For ARIMA, optimal orders vary across stocks because each stock exhibits different autocorrelation and partial autocorrelation structure; the most common selections were $(p,d,q) = (2,1,2)$ and $(1,1,1)$. XGBoost uses early stopping with a patience of 50 rounds on validation RMSE, so the effective number of estimators is often lower than the specified maximum. The LSTM and Simple Transformer sequence lengths are fixed at 252 trading days (one calendar year) to match the proposed model's input window, ensuring that differences in performance reflect architectural capacity rather than information asymmetry. The MS-VAR uses the MSIH specification (Markov-switching intercept and heteroscedasticity), which allows both the intercept and the error variance to switch across regimes while keeping the autoregressive coefficients regime-invariant; this provides sufficient flexibility to capture volatility regime changes without overfitting the transition dynamics. For the HMM-LSTM, regime assignments are determined by the Viterbi path on the training set, and each regime-specific LSTM is trained only on data segments assigned to its corresponding state. PatchTST and iTransformer use the channel-independent and inverted-attention configurations recommended in their respective original publications, with sequence lengths fixed at 252 to match other deep learning baselines.

%========================================
% REFERENCES
%========================================
\bibliographystyle{IEEEtran}
\bibliography{references}

@article{fischer2018deep,
  title={Deep learning with long short-term memory networks for financial market predictions},
  author={Fischer, Thomas and Krauss, Christopher},
  journal={European Journal of Operational Research},
  volume={270},
  number={2},
  pages={654--669},
  year={2018},
  doi={10.1016/j.ejor.2017.11.054}
}

@inproceedings{vaswani2017attention,
  title={Attention is all you need},
  author={Vaswani, Ashish and Shazeer, Noam and Parmar, Niki and Uszkoreit, Jakob and Jones, Llion and Gomez, Aidan N and Kaiser, {\L}ukasz and Polosukhin, Illia},
  booktitle={Advances in Neural Information Processing Systems},
  volume={30},
  pages={5998--6008},
  year={2017}
}

@article{wu2020comprehensive,
  title={A comprehensive survey on graph neural networks},
  author={Wu, Zonghan and Pan, Shirui and Chen, Fengwen and Long, Guodong and Zhang, Chengqi and Philip, S Yu},
  journal={IEEE Transactions on Neural Networks and Learning Systems},
  volume={32},
  number={1},
  pages={4--24},
  year={2020},
  doi={10.1109/tnnls.2020.2978386}
}

@inproceedings{wu2022nodeformer,
  title={NodeFormer: A scalable graph structure learning transformer for node classification},
  author={Wu, Qitian and Zhao, Wentao and Li, Zenan and Wipf, David P and Yan, Junchi},
  booktitle={Advances in Neural Information Processing Systems},
  volume={35},
  pages={27387--27401},
  year={2022}
}

@inproceedings{devlin2019bert,
  title={{BERT}: Pre-training of deep bidirectional transformers for language understanding},
  author={Devlin, Jacob and Chang, Ming-Wei and Lee, Kenton and Toutanova, Kristina},
  booktitle={Proceedings of the Conference of the North American Chapter of the Association for Computational Linguistics},
  pages={4171--4186},
  year={2019},
  doi={10.18653/v1/N19-1423}
}

@inproceedings{liu2022bilstm,
  title={A new deep network model for stock price prediction},
  author={Liu, Mingchen and Sheng, Haoliang and Zhang, Nan and others},
  booktitle={International Conference on Machine Learning for Cyber Security},
  pages={413--426},
  year={2022},
  doi={10.1007/978-3-031-20102-8_32}
}

@article{chen2021graph,
  title={A novel graph convolutional feature based convolutional neural network for stock trend prediction},
  author={Chen, Wei and Jiang, Manyi and Zhang, Wei-Guo and Chen, Zhensong},
  journal={Information Sciences},
  volume={556},
  pages={67--94},
  year={2021},
  doi={10.1016/j.ins.2020.12.068}
}

@article{wang2022multigraph,
  title={{MG-Conv}: A spatiotemporal multi-graph convolutional neural network for stock market index trend prediction},
  author={Wang, Chao and Liang, Haibo and Wang, Bin and Cui, Xiaomeng and Xu, Yi},
  journal={Computers and Electrical Engineering},
  volume={103},
  pages={108285},
  year={2022},
  doi={10.1016/j.compeleceng.2022.108285}
}

@article{alridhawi2025nodeformer,
  title={Stock Market Prediction Using Node Transformer Architecture Integrated with {BERT} Sentiment Analysis},
  author={Al Ridhawi, Mohammad A and Haj Ali, Mahtab and Al Osman, Hussein},
  journal={Submitted to IEEE Access},
  year={2026},
  note={Under review. arXiv:2603.05917}
}

@article{hamilton1989new,
  title={A new approach to the economic analysis of nonstationary time series and the business cycle},
  author={Hamilton, James D},
  journal={Econometrica},
  volume={57},
  number={2},
  pages={357--384},
  year={1989},
  doi={10.2307/1912559}
}

@article{diebold1994regime,
  title={Regime switching with time-varying transition probabilities},
  author={Diebold, Francis X and Lee, Joon-Haeng and Weinbach, Gretchen C},
  journal={Business Cycles: Durations, Dynamics, and Forecasting},
  pages={144--165},
  year={1994},
  publisher={Princeton University Press}
}

@book{tong1990non,
  title={Non-linear time series: A dynamical system approach},
  author={Tong, Howell},
  year={1990},
  publisher={Oxford University Press}
}

@article{nystrup2017clustering,
  title={Regime-based versus static asset allocation: Letting the data speak},
  author={Nystrup, Peter and Madsen, Henrik and Lindstr{\"o}m, Erik},
  journal={The Journal of Portfolio Management},
  volume={44},
  number={1},
  pages={103--115},
  year={2017},
  doi={10.3905/jpm.2015.42.1.103}
}

@article{aminikhanghahi2017survey,
  title={A survey of methods for time series change point detection},
  author={Aminikhanghahi, Samaneh and Cook, Diane J},
  journal={Knowledge and Information Systems},
  volume={51},
  number={2},
  pages={339--367},
  year={2017},
  doi={10.1007/s10115-016-0987-z}
}

@article{hinton2006reducing,
  title={Reducing the dimensionality of data with neural networks},
  author={Hinton, Geoffrey E and Salakhutdinov, Ruslan R},
  journal={Science},
  volume={313},
  number={5786},
  pages={504--507},
  year={2006},
  doi={10.1126/science.1127647}
}

@inproceedings{kingma2014auto,
  title={Auto-encoding variational {B}ayes},
  author={Kingma, Diederik P and Welling, Max},
  booktitle={International Conference on Learning Representations},
  year={2014}
}

@article{pumsirirat2018credit,
  title={Credit card fraud detection using deep learning based on auto-encoder and restricted {B}oltzmann machine},
  author={Pumsirirat, Apapan and Yan, Liu},
  journal={International Journal of Advanced Computer Science and Applications},
  volume={9},
  number={1},
  pages={18--25},
  year={2018},
  doi={10.14569/ijacsa.2018.090103}
}

@article{ahmed2016survey,
  title={A survey of network anomaly detection techniques},
  author={Ahmed, Mohiuddin and Mahmood, Abdun Naser and Hu, Jiankun},
  journal={Journal of Network and Computer Applications},
  volume={60},
  pages={19--31},
  year={2016},
  doi={10.1016/j.jnca.2015.11.016}
}

@article{bao2017deep,
  title={A deep learning framework for financial time series using stacked autoencoders and long-short term memory},
  author={Bao, Wei and Yue, Jun and Rao, Yulei},
  journal={PLOS ONE},
  volume={12},
  number={7},
  pages={e0180944},
  year={2017},
  doi={10.1371/journal.pone.0180944}
}

@book{sutton2018reinforcement,
  title={Reinforcement learning: An introduction},
  author={Sutton, Richard S and Barto, Andrew G},
  year={2018},
  edition={2nd},
  publisher={MIT Press}
}

@article{jiang2017deep,
  title={A deep reinforcement learning framework for the financial portfolio management problem},
  author={Jiang, Zhengyao and Xu, Dixing and Liang, Jinjun},
  journal={arXiv preprint arXiv:1706.10059},
  year={2017}
}

@article{ning2021double,
  title={A double deep {Q}-learning model for energy-efficient edge scheduling},
  author={Ning, Zhaolong and Dong, Peiran and Wang, Xiaojie and Hu, Xiping and Guo, Lei and Hu, Bin and Kwok, Ricky Y K and Leung, Victor C M},
  journal={IEEE Transactions on Services Computing},
  volume={14},
  number={5},
  pages={1555--1566},
  year={2021},
  doi={10.1109/tsc.2018.2867482}
}

@article{deng2016deep,
  title={Deep direct reinforcement learning for financial signal representation and trading},
  author={Deng, Yue and Bao, Feng and Kong, Youyong and Ren, Zhiquan and Dai, Qionghai},
  journal={IEEE Transactions on Neural Networks and Learning Systems},
  volume={28},
  number={3},
  pages={653--664},
  year={2016},
  doi={10.1109/tnnls.2016.2522401}
}

@article{mnih2015human,
  title={Human-level control through deep reinforcement learning},
  author={Mnih, Volodymyr and Kavukcuoglu, Koray and Silver, David and Rusu, Andrei A and Veness, Joel and Bellemare, Marc G and Graves, Alex and Riedmiller, Martin and Fidjeland, Andreas K and Ostrovski, Georg and others},
  journal={Nature},
  volume={518},
  number={7540},
  pages={529--533},
  year={2015},
  doi={10.1038/nature14236}
}

@inproceedings{haarnoja2018soft,
  title={Soft actor-critic: Off-policy maximum entropy deep reinforcement learning with a stochastic actor},
  author={Haarnoja, Tuomas and Zhou, Aurick and Abbeel, Pieter and Levine, Sergey},
  booktitle={International Conference on Machine Learning},
  pages={1861--1870},
  year={2018},
  organization={PMLR}
}

@article{ang2002international,
  title={International asset allocation with regime shifts},
  author={Ang, Andrew and Bekaert, Geert},
  journal={The Review of Financial Studies},
  volume={15},
  number={4},
  pages={1137--1187},
  year={2002},
  doi={10.1093/rfs/15.4.1137}
}

@article{guidolin2007asset,
  title={Asset allocation under multivariate regime switching},
  author={Guidolin, Massimo and Timmermann, Allan},
  journal={Journal of Economic Dynamics and Control},
  volume={31},
  number={11},
  pages={3503--3544},
  year={2007},
  doi={10.1016/j.jedc.2006.12.004}
}

@article{haarnoja2019sac,
  title={Soft actor-critic algorithms and applications},
  author={Haarnoja, Tuomas and Zhou, Aurick and Hartikainen, Kristian and Tucker, George and Ha, Sehoon and Tan, Jie and Kumar, Vikash and Zhu, Henry and Gupta, Abhishek and Abbeel, Pieter and Levine, Sergey},
  journal={arXiv preprint arXiv:1812.05905},
  year={2019}
}

@article{liu2022vix,
  title={{VIX} and stock market volatility predictability: A new approach},
  author={Liu, Zhichao and Liu, Jing and Zeng, Qing and Wu, Lan},
  journal={Finance Research Letters},
  volume={48},
  pages={102887},
  year={2022},
  publisher={Elsevier},
  doi={10.1016/j.frl.2022.102887}
}

@inproceedings{cortis2017semeval,
  title={{SemEval}-2017 Task 5: Fine-grained sentiment analysis on financial microblogs and news},
  author={Cortis, Keith and Freitas, Andr{\'e} and Daudert, Tobias and Huerlimann, Manuela and Zarrouk, Manel and Handschuh, Siegfried and Davis, Brian},
  booktitle={Proceedings of the 11th International Workshop on Semantic Evaluation (SemEval-2017)},
  pages={519--535},
  year={2017}
}

@inproceedings{nie2023patchtst,
  title={A Time Series is Worth 64 Words: Long-term Forecasting with Transformers},
  author={Nie, Yuqi and Nguyen, Nam H. and Sinthong, Phanwadee and Kalagnanam, Jayant},
  booktitle={Proceedings of the 11th International Conference on Learning Representations (ICLR)},
  year={2023}
}

@inproceedings{liu2024itransformer,
  title={{iTransformer}: Inverted Transformers Are Effective for Time Series Forecasting},
  author={Liu, Yong and Hu, Tengge and Zhang, Haoran and Wu, Haixu and Wang, Shiyu and Ma, Lintao and Long, Mingsheng},
  booktitle={Proceedings of the 12th International Conference on Learning Representations (ICLR)},
  year={2024}
}

@inproceedings{wu2023timesnet,
  title={{TimesNet}: Temporal 2D-Variation Modeling for General Time Series Analysis},
  author={Wu, Haixu and Hu, Tengge and Liu, Yong and Zhou, Hang and Wang, Jianmin and Long, Mingsheng},
  booktitle={Proceedings of the 11th International Conference on Learning Representations (ICLR)},
  year={2023}
}

@book{krolzig1997markov,
  title={Markov-Switching Vector Autoregressions: Modelling, Statistical Inference, and Application to Business Cycle Analysis},
  author={Krolzig, Hans-Martin},
  year={1997},
  publisher={Springer-Verlag},
  address={Berlin}
}

@article{diebold1995comparing,
  title={Comparing Predictive Accuracy},
  author={Diebold, Francis X. and Mariano, Roberto S.},
  journal={Journal of Business \& Economic Statistics},
  volume={13},
  number={3},
  pages={253--263},
  year={1995},
  doi={10.1080/07350015.1995.10524599}
}

@book{box1976time,
  title={Time Series Analysis: Forecasting and Control},
  author={Box, George E. P. and Jenkins, Gwilym M.},
  year={1976},
  publisher={Holden-Day},
  address={San Francisco}
}

@article{sims1980macroeconomics,
  title={Macroeconomics and Reality},
  author={Sims, Christopher A.},
  journal={Econometrica},
  volume={48},
  number={1},
  pages={1--48},
  year={1980},
  doi={10.2307/1912017}
}

@article{breiman2001random,
  title={Random Forests},
  author={Breiman, Leo},
  journal={Machine Learning},
  volume={45},
  number={1},
  pages={5--32},
  year={2001},
  doi={10.1023/A:1010933404324}
}

@book{vapnik1995nature,
  title={The Nature of Statistical Learning Theory},
  author={Vapnik, Vladimir N.},
  year={1995},
  publisher={Springer},
  address={New York},
  doi={10.1007/978-1-4757-2440-0}
}

@inproceedings{chen2016xgboost,
  title={{XGBoost}: A Scalable Tree Boosting System},
  author={Chen, Tianqi and Guestrin, Carlos},
  booktitle={Proceedings of the 22nd ACM SIGKDD International Conference on Knowledge Discovery and Data Mining},
  pages={785--794},
  year={2016},
  doi={10.1145/2939672.2939785}
}

@article{hochreiter1997long,
  title={Long Short-Term Memory},
  author={Hochreiter, Sepp and Schmidhuber, J{\"u}rgen},
  journal={Neural Computation},
  volume={9},
  number={8},
  pages={1735--1780},
  year={1997},
  doi={10.1162/neco.1997.9.8.1735}
}

\begin{IEEEbiography}[{\includegraphics[width=1in,height=1.25in,clip,keepaspectratio]{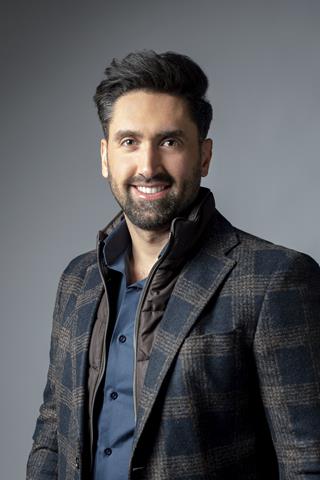}}]{Mohammad Al Ridhawi}
received the B.A.Sc.\ degree in computer engineering and the M.Sc.\ degree in digital transformation and innovation (machine learning) from the University of Ottawa, Ottawa, Canada, in 2019 and 2021, respectively. He is currently pursuing the Ph.D.\ degree in electrical and computer engineering at the University of Ottawa, where he also serves as a Part-Time Engineering Professor. He has industry experience as a Senior Data Scientist and Senior Machine Learning Engineer, building production ML systems in financial and environmental domains. His research interests include deep learning, graph neural networks, natural language processing, financial time series analysis, and reinforcement learning.
\end{IEEEbiography}

\begin{IEEEbiography}[{\includegraphics[width=1in,height=1.25in,clip,keepaspectratio]{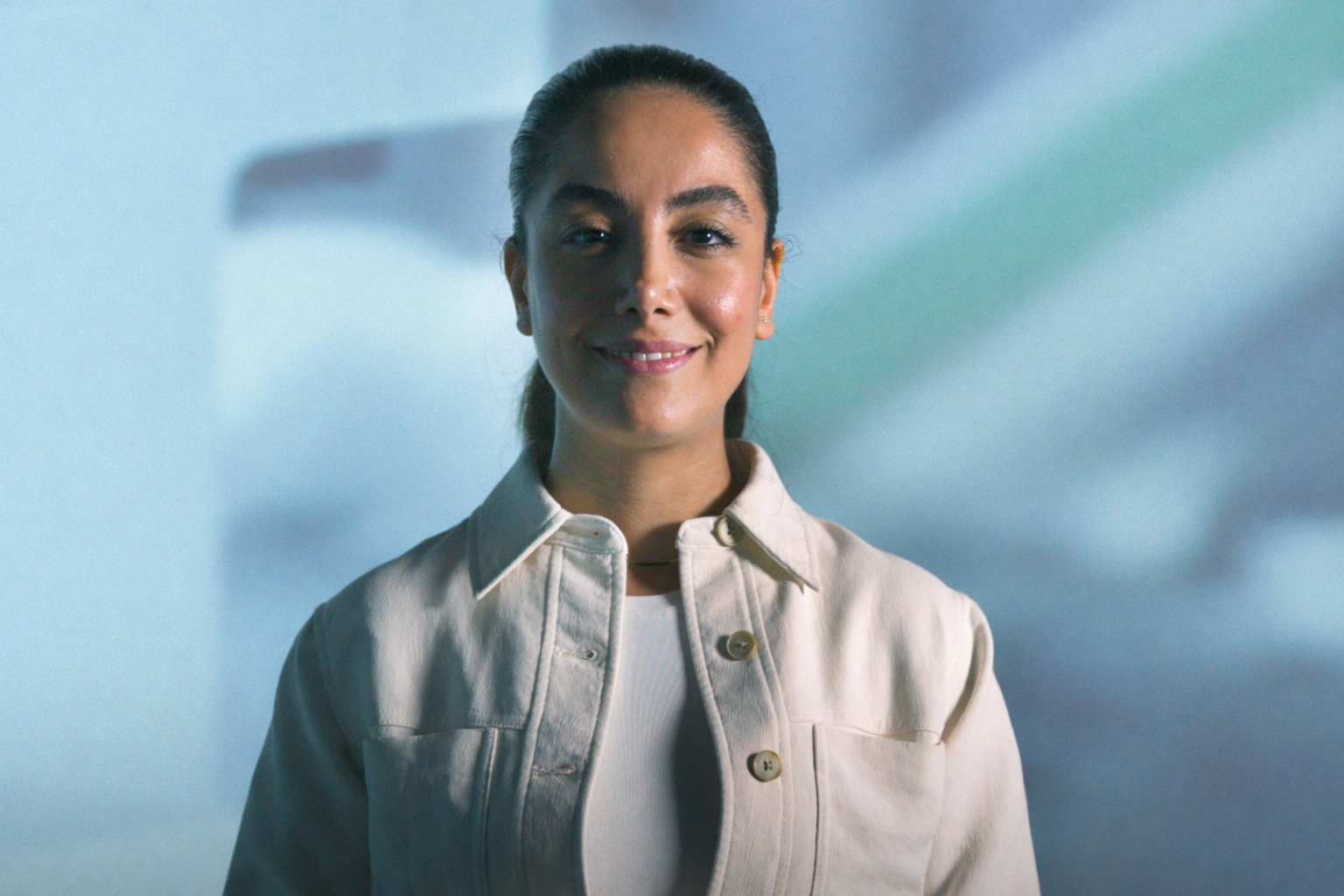}}]{Mahtab Haj Ali}
received the M.Sc.\ degree in digital transformation and innovation from the University of Ottawa, Ottawa, Canada, in 2021. She is currently pursuing the Ph.D.\ degree in electrical and computer engineering at the University of Ottawa, with a research focus on time series forecasting and deep learning models. She works as an AI Research Engineer at the National Research Council of Canada, where she builds and evaluates large language models (LLMs) and develops AI-driven solutions for real-world industrial applications. Her work includes large-scale time series analysis, advanced feature engineering, and the application of LLMs in production environments. Her research interests include deep learning for time series analysis, deep neural networks, and applied artificial intelligence.
\end{IEEEbiography}

\begin{IEEEbiography}[{\includegraphics[width=1in,height=1.25in,clip,keepaspectratio]{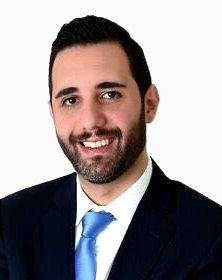}}]{Hussein Al Osman}
received the B.A.Sc., M.A.Sc., and Ph.D.\ degrees from the University of Ottawa, Ottawa, Canada. He is an Associate Professor and Associate Director in the School of Electrical Engineering and Computer Science at the University of Ottawa, where he leads the Multimedia Processing and Interaction Group. His research focuses on affective computing, multimodal affect estimation, human--computer interaction, serious gaming, and multimedia systems. He has produced over 50 peer-reviewed research articles, two patents, and several technology transfers to industry.
\end{IEEEbiography}

\EOD

\end{document}